\documentclass[final,5p,times,twocolumn]{elsarticle}

\makeatletter
\def\ps@pprintTitle{%
   \let\@oddhead\@empty
   \let\@evenhead\@empty
   \let\@oddfoot\@empty
   \let\@evenfoot\@oddfoot
}
\makeatother

\usepackage{tabularx,booktabs}
\usepackage[version=4]{mhchem}
\usepackage{siunitx}
\usepackage{ragged2e}
\usepackage{rotating}
\usepackage{textcomp}  
\usepackage{longtable, tabularx, rotating, booktabs, ragged2e}
\usepackage{nameref,hyperref}

\usepackage{amssymb}
\usepackage{amsmath}

\begin{document}

\begin{frontmatter}



\title{A Guide for Manual Annotation of Scientific Imagery: How to Prepare for Large Projects}

\author[label1]{Azim Ahmadzadeh}
\ead{ahmadzadeh@umsl.edu}
\author[label2]{Rohan Adhyapak}
\author[label2]{Armin Iraji}
\author[label2]{Kartik Chaurasiya}
\author[label3]{V Aparna}
\author[label2]{Petrus C. Martens}

\affiliation[label1]{organization={University of Missouri - St. Louis},
            }
\affiliation[label2]{organization={Georgia State University}}
\affiliation[label3]{organization={Bay Area Environmental Research Institute}}

\begin{abstract}
    Despite the high demand for manually annotated image data, managing complex and costly annotation projects remains under-discussed. This is partly due to the fact that leading such projects requires dealing with a set of diverse and interconnected challenges which often fall outside the expertise of specific domain experts, leaving practical guidelines scarce. These challenges range widely from data collection to resource allocation and recruitment, from mitigation of biases to effective training of the annotators. This paper provides a domain-agnostic preparation guide for annotation projects, with a focus on scientific imagery. Drawing from the authors' extensive experience in managing a large manual annotation project, it addresses fundamental concepts including success measures, annotation subjects, project goals, data availability, and essential team roles. Additionally, it discusses various human biases and recommends tools and technologies to improve annotation quality and efficiency. The goal is to encourage further research and frameworks for creating a comprehensive knowledge base to reduce the costs of manual annotation projects across various fields.
\end{abstract}





\end{frontmatter}


\section{Introduction}
    Computer vision, with the emergence of deep neural networks and the much needed computing power in hardware devices, has made impressive advances during the past decade \citep{lecun2019deep}. As it is often said, AI is driven by data, not code, the quality and quantity in such datasets set the limits of today's computer vision capabilities. Some of the most influential image datasets---outdated or not---are MNIST \citep{lecun1998mnist}, BSDS \citep{martin2001database}, LabelMe \citep{russell2008labelme}, ImageNet \citep{deng2009imagenet}---the most popular image dataset with bounding-box annotations---CIFAR \citep{krizhevsky2009learning}, PASCAL VOC \citep{everingham2010pascal}, MS COCO \cite{lin2014coco}---the most popular image dataset with segmentation annotation---Places \citep{zhou2014learning}, and LSUN \citep{yu2015lsun}. With the rapid advances in computer vision algorithms, newer datasets are created to bring new challenges for the algorithms.

    
    Despite the growing demands of manual annotation of visual data, there have not yet been many efforts for sharing the experience of managing the annotation task. Published studies releasing new datasets often limit their papers to the machine learning technical side that is deemed important for the future model designers as well as the contestants of the competitions. This is partially done due to constraints such as scope and page limits enforced by publishers. For example, consider LVIS, an impressively large image dataset ($\approx$2M instances) created by Facebook AI Research team, with over 1,000 categories of objects \citep{Gupta2019lvis}. While the paper (and its extended version available on arXiv) goes into sufficient depth useful for future model developers, there are few details regarding how the actual segmentation task is done by the annotators. Other studies---mostly focusing on reducing the noise in annotation of very large image datasets---discuss improving the annotation efficiency by using machine learning models next to human annotators (see \cite{Branson2017lean, Liao2021towards} and references therein). However, leading such large projects requires handling a wide spectrum of challenges, from developing software products for data collection, to engineering pipelines for making the manual annotation possible, to resource management, and dealing with potential human and machine biases throughout the entire process. Such challenges are sometimes so deeply interconnected with each other, as well as with those of the subject matter, that holistic approaches may be more efficient than ad hoc treatments suitable for individual projects. Moreover, manual annotation of data is often a tedious and time-consuming task which makes it costly, demanding efficiency and methods to minimize the risks. As this study aims to show, for the manual annotation of scientific objects/events (as opposed to common objects), many of such challenges require unique strategies due to their fundamentally different objectives.

    For the manual annotation of (very) large datasets, the common approach is to use a crowd-sourcing technology, such as the Amazon Mechanical Turk (MTurk) (as an example, see \cite{Gupta2019lvis}). In fact, one could argue that without such platforms, it may be impractical to annotate millions of objects within a reasonable time frame. However, the `` `Turk philosophy' (human computation is cheap and subsumes automated methods)'' does not always hold \citep{vondrick2013efficiently}, and even less so for complex scientific objects/events. There is also the ethical aspect of using such platforms. There has been a growing awareness regarding some exploitative aspects involved in such practices that we encourage the reader to consult the literature, for instance, see \cite{dolmaya2021ethics, moss2023ethical, standing2018ethical, Hao_2023,Velasco_2024}.
       
    The main idea behind such a solution is that an annotation pipeline can be engineered such that the human annotators only need to follow a short list of very simple rules, bringing down the costs by reducing (or completely eliminating) the need for their involvement in the actual project. To have a better understanding of the situation, let us consider an example. In the annotation protocol introduced by \cite{su2012crowdsourcing}, there were only four rules for the annotators to draw bounding-boxes around objects: (1) each object should be fully enclosed within a bounding-box, (2) no object may be missed, (3) objects should be annotated one at a time, and (4) the annotation of an image is considered complete if no object in that image is left without a bounding-box. Although the rules are simple, even for drawing bounding-boxes (which is one of the least complex types of annotations), the annotation pipeline requires that each annotator pass a ``qualification test''. It also sends each object to another annotator for independent validation. There is a set of three rules for the reviewers to follow describing what a ``good bounding-box'' looks like. It should not be difficult to see that even such a simple task requires a complex system to help the annotators do their best. As reported in the same study, with their efficient protocols, the bounding-box annotation of an object takes $88.0s$, on average (with a median of $42.4s$), when tested on a subset of the ImageNet dataset \citep{deng2009imagenet, russakovsky2015imagenet}. It should be noted that the annotators were asked to annotate a specific type of object per task, which was a time saver. For example, for an image presented on the screen, the text prompt reads ``Draw a box around balloon''. Therefore, the annotators did not spend any time searching the image for any of the many objects listed in a separate list (i.e., ultimately, 1000 categories). The time needed to identify that at least one balloon was present in the image should be added to the annotation time. For a more efficient bounding-box annotation method, and a thorough literature review of annotation time and efficiency, we refer the reader to \cite{papadopoulos2017extreme}.
        
    The manual annotation of image data is indeed a laborious task. Adding the scientific objects/events to the equation brings an entirely new set of challenges to the table. When dealing with scientific objects/events that are unfamiliar to the annotators, the definitions of classes can be complex and may require trained eyes. Despite its unique challenges, because such datasets are usually much smaller than those used for the common object-detection challenges, the process is less frequently documented, or it is heavily domain-specific and therefore contains little transferable knowledge. In this paper, we present a guide for manual annotation of scientific objects/events without targeting any specific domain. It should be emphasized, however, that this is not an exhaustive study, but rather a starting point for similar studies and technical reports to build upon ours, resulting in more effective data annotation pipelines, higher-quality annotated datasets, and consequently, more robust machine learning models to be used in various disciplines of science and engineering. The content of this paper is mostly based on the authors' first-hand experience in manual annotation and management of a large annotation project. The technologies and services listed should be taken only as examples, and the authors do not intend to express any preference for any commercial or non-commercial product. 
    
\section{Measures of Success}\label{sec:success}
    In this study, we define the success of an image-dataset annotation task by the following domain-agnostic criteria: \textbf{C1.} The pre-defined annotation metadata are produced as expected in quality and quantity. \textbf{C2.} The pre-defined annotation metadata are produced as expected in shape, format, granularity, etc. \textbf{C3.} The pre-defined annotation metadata are produced without a significant increase in the needed resources. \textbf{C4.} The pre-defined end goals are fully achieved using the produced annotation metadata. \textbf{C5.} The entire project is fully documented for both the annotation team and the domain experts.

    The first criterion (C1) is ideally regarded as the only criterion for an annotation project, which speaks to its importance. Here, the term `quality' concerns the precision and accuracy of the annotations. For example, minimum bounding rectangles (MBR) are supposed to tightly enclose objects in images. If many of the created MBRs only loosely enclose objects, the first criterion is not fully met. The term `quantity' concerns the expected number of objects or images to be annotated. Often, compromising the quality and/or quantity of the annotation products is the first target in an over-budget project. Moreover, since discrepancies in quantity are generally easier to spot than those in quality, the quality factors are more likely to be compromised. This is why having a set of quantifiable metrics (if possible) for measuring the quality of the final products is critical.
    
    The second criterion (C2) targets the annotation geometry. Occasionally, a specific type of annotation (e.g., line, curve, rectangle, polygon, mask, etc.) may be of great value for domain experts, yet it may be impractical to expect the annotators to produce them. This could happen due to the fact that the expected metadata might be more time-consuming to produce than annotators are compensated for, or it might be beyond the knowledge of the annotators, or even beyond the capabilities of the chosen annotation platform. For example, the team may initially decide to produce segmentation masks for objects (to have pixel-precise annotation capability), but the annotation platform may only offer polygons (whose granularities are limited by the number of corner points the annotators create for each object). Therefore, it is critical for the team to assess the feasibility of what domain experts might demand, and investigate ways to meet their requirements, if possible.

    The third criterion (C3) is a typical management concern, i.e., resource management. While underestimating the project duration and budget is simply a hard reality that happens more frequently than desired, it must be noted that in some settings, re-budgeting or extending the project timeline is simply not an option. For example, in an academic setting where the funding comes from an external source for a proposed project, it is not common to request extra funds. This may easily result in wrapping up the project without meeting all the basic requirements, or even abandoning the project altogether. In other sectors where extra funds may be requested, extending the final deadline may be a bigger issue as the annotation project is often a piece in a much more time-sensitive production domino.

    The fourth criterion (C4) reminds the annotation team that the usability of their product is entirely their responsibility. The lack of sufficient details about the end goals is the primary reason for producing ineffective or entirely useless annotated data, despite meeting all the technically agreed-upon requirements. The main cause of this is a broken communication channel between the annotators and the domain experts. The team may identify this significant issue during the annotation process, however, it could be too late and costly to overhaul the project by then. Therefore, there is a temptation to proceed with the existing plan, hoping for a technical miracle.

    The fifth criterion (C5) addresses the transparency of the project. In any annotation project, there are numerous decisions made, both at the microscopic level (e.g., specific instructions given to the annotators for disregarding ``unimportant'' objects) and at the macroscopic level (e.g., strategies for maintaining an unbiased annotation pipeline). Those decisions and how they were carried out throughout the project provide context for the annotated data, exposing the strengths and the weaknesses to end users. Such a detailed documentation is analogous to a warranty note of a product or service, as they both help end users understand which features of the data they can and cannot rely on. For example, suppose an annotation team permits the annotators to modify the image color distribution (such as contrast) in an attempt to aid them in discerning details in order to produce high-quality annotations. In that case, the end users should be warned that for highly noisy images, the background noise could occasionally be mistaken by the annotators for parts of the foreground objects. Such mistakes could then result in the formation of unrealistic structures. The lack of such details in the documentation of the annotation project may leave end users to trust the segmentations without taking into account the impact of noisy images. In contrast, knowing about this decision allows end users to consider removing segmentations corresponding to noisy images, resulting in more reliable operational models and scientific discoveries. Furthermore, a comprehensive documentation transforms the end product from a mere dataset into something more substantial; a dynamic entity which can be further refined in future iterations, whether by the same team or a different one.

\section{Subjects of Study and End Goals}\label{sec:subject}
    The subjects of study and the end goals should not be considered as common knowledge to be overlooked. There are many details which should be thoroughly discussed and agreed upon. Those details play an essential role in forming the numerous micro decisions which collectively determine the success of an annotation project. Below, we list the details regarding the subjects of study and the end goals of annotation, which should be carefully examined before even the planning phase begins.
    
    \subsection{Objects/Events of Interest (OoIs)}\label{subsec:OoIs}
\begin{figure*}[t]
            \includegraphics[width=1.0\linewidth]{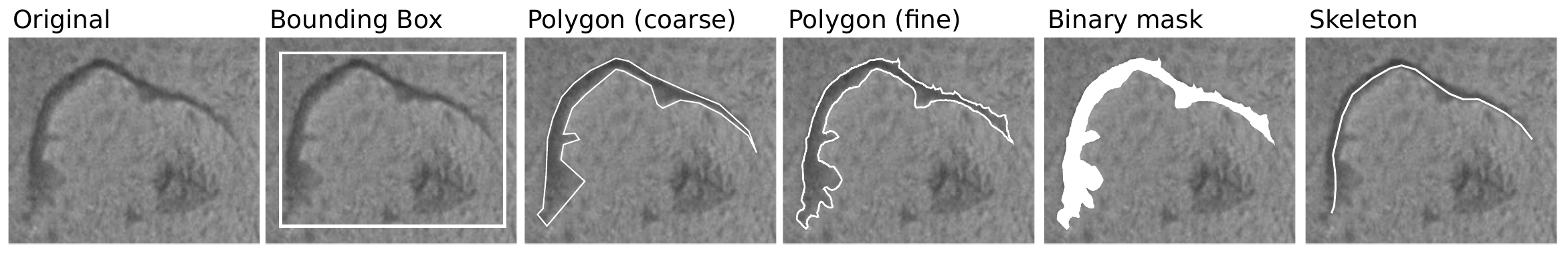}
            \centering
            \caption{The figure shows different types of the popular PoIs overlaid on top of an instance of a solar filament as an example. From left to right, it shows: the original OoI, a bounding-box, a polygon with low granularity, a polygon with high granularity, a binary mask, and a skeleton. In this illustration, the skeleton PoI is used to capture the `spine' of the filament which is of value for heliophysicists interested in the chirality of the magnetic field at that region. The filament is observed by the Global Oscillation Network Group on 2023-07-21 at 18:44:22 UT.}
            \label{fig:types_of_PoIs}
        \end{figure*}

        An \textit{Object/Event of Interest}, OoI in short, is an object or event which needs to be annotated. This is also called the \textit{class}, \textit{label}, or \textit{category} of instances. For example, in the Microsoft Common Objects in Context (COCO) dataset---one of the most popular datasets of common objects---there are over half a million instances of 80 different OoIs, such as `person', `knife', `chair', `book', etc. \citep{lin2014coco}. To give a more task-specific example, in a defect detection study on printed electronic circuit boards, the team identified the various types of defects on electronic boards, i.e., their OoIs, as `exposed copper', `broken hole', `line damage', `scratch', `pinhole', `short', and `open' \citep{lue2021fpcb}. We use the term `OoI instance' to refer to an instance of an OoI. For example, Fig. \ref{fig:types_of_PoIs} shows a solar filament as an OoI instance annotated using different methods.
        
        It is crucial that the team spend enough time studying all OoIs whose annotation is necessary to serve the end goals of the project. That said, there is a trade-off in the number of OoIs to be identified. On one hand, having a short list of OoIs comes with a greater chance of missing an important OoI. Since the design of the annotation workflow, the training phase, and the evaluation process, all heavily depend on what OoIs are expected to be identified, adding a new OoI when all those pieces are already developed can be costly. Not including this new OoI, needless to say, takes away the chance of building a representative dataset. On the other hand, the inclusion of a large set of OoIs comes with the risk of redundancy, which is also costly. More OoIs equal more manual labor. To give the reader a tangible sense of the extra cost, consider the COCO dataset again. As reported in \cite{lin2014coco}, the team had to spend roughly 22,000 worker hours only to identify the presence of one instance of any of the 11 super-categories in each image before beginning the annotation of the 80 unique OoIs. Adding any extra OoI to the list obviously burdens the resources: including more OoIs simply makes the annotation task more difficult for the human annotators. The implication of this is often hidden from the management team. Harder tasks require longer training sessions. More importantly, the harder the tasks are the more difficult it gets to keep the annotators who are already trained and experienced. The management team should consider either a higher compensation rate, or the extra cost of more frequent recruitment and training. Each of those decisions stretches the allocated resources thinner.
        
        The OoIs list should also be exclusive in terms of what the annotators may need when they cannot confidently assign an OoI to an instance. For this, the team can introduce an auxiliary OoI for catching the less trivial cases. Those OoIs can be named `in-between', `not-defined', or `uncertain', to give a few examples. Of course, this might be redundant in projects where the distinction between the OoIs is believed to be trivial. But in the context of scientific objects (e.g., microorganisms in microscopic imagery), even domain experts may not always be able to confidently classify an instance. This could be the case for numerous reasons, to name a few: (1) an instance's features may fall below the resolution of the image, (2) the background noise may obscure the key features of an instance, or (3) the literature's definition of an OoI might not sufficiently encompass all its various manifestations in the collected observations. It is therefore expected that some instances exhibit the key features of two or more OoIs. The absence of an auxiliary OoI to catch such non-trivial instances may occasionally force the annotators to randomly choose one OoI over another, or let their bias take over, leading to some artificial distinctions. (see Sec.~\ref{sec:bias} for a discussion on the impacts of bias in annotation). Such errors are very difficult to be identified and corrected later on. Introducing a separate OoI to catch such instances will increase the overall reliability and consistency of the annotated data. Moreover, this auxiliary OoI speeds up the manual annotation by making the decision-making process more natural. Consequently, it reduces the mental fatigue of the annotators, caused by having to constantly deal with the dilemma of the classification of instances which equally match the properties of more than one OoI.

    \subsection{Properties of Interest (PoIs)}
        
        An OoI refers to an object or event that should be annotated, but it does not specify what properties should be annotated and how. A \textit{Property of Interest}, in short, PoI, refers to a specific property which needs to be captured for all/some OoIs during annotation. If no PoI is specified, the annotators are expected to perform only whole-image labeling (i.e., image classification). This task involves assigning one or more of the given OoIs to an image indicating the presence of instances of the corresponding OoIs in that image. No other information will be stored. In this section, we review some of the most popular PoIs, as shown in Fig.~\ref{fig:types_of_PoIs}.

        The simplest PoI to create is \textit{coordinates}. Coordinates of an instance locates the center of that instance. How the `center' is defined depends on the task. Sometimes, this is only an intermediary step during a multi-stage annotation process and may not hold much value for the training of an object detection algorithm. In such cases, the center is often not meant literally (as used in \cite{lin2014coco}, called instance spotting), and any point roughly in the middle of the area occupied by each OoI instance qualifies.
        
        A simple yet much more useful PoI is called a \textit{bounding-box} (bbox). A bounding-box is a rectangle confining the entire area of each instance of OoI. Usually, a bounding-box is implied to be axis-aligned, meaning that its edges are parallel to the Cartesian coordinate axes---otherwise it is just a polygon with four edges. Whether the box is a \textit{minimum bounding rectangle} (MBR) or just a fairly tight box should be defined by the team. An MBR is often defined as the smallest axis-aligned rectangle (or, in higher dimensions, a hyperrectangle) that completely encloses an OoI's instance. A bounding-box gives a good estimate of the size of an instance (not so good for narrow shapes) as well as its location. ImageNet, another very popular image dataset used for object detection competitions for seven years (2010 - 2017) \citep{russakovsky2015imagenet}, comes with the bounding-boxes of instances of 200 unique OoIs identified in 1.2 million images.

        A polygon (a closed path) is another PoI which can be used to capture the shape, size, and location of an object more precisely than a bbox. Although polygons have their own use cases, they give their place to the segmentations (discussed next) whenever the structural details in the OoIs become important. This is simply because for capturing a detailed shape the annotator must create and manipulate too many points, one point at a time.
        
        To obtain a detailed boundary of each instance, and to capture its shape, size, and location, \textit{segmentation} may be considered. A segmentation is a collection of pixels over which an instance of OoI is spanned. There are two types of segmentations: (1) \textit{semantic segmentation} where each pixel in an image is assigned to an OoI (e.g., all cars in an image are considered as one instance of the `car' OoI), and (2) \textit{instance segmentation} where each pixel in an image is assigned to an instance of an OoI. In other words, an instance, regardless of its corresponding OoI, is recognized individually (e.g., each car of several cars visible in an image is annotated as a unique instance of the `car' OoI). The Open Images Dataset V7 (created by Google) is the most recently used dataset of everyday objects, that contains segmentation of over 2.7 million instances of 350 OoIs \citep{openimagedataset2022google}. This dataset has been the subject of the competitions in the Robust Vision Challenge since 2018 \citep{kuznetsova2020open}. 
        
        A segmentation PoI is typically either encoded as a polygon or a binary mask, specifying the region in which an OoI instance is confined. While the choice between these two representations might be dictated by the annotation platform of choice and how rich its toolbox is, some considerations should be noted. A polygon is a (relatively) low-dimensional data point, i.e., an array of as many coordinates as the number of vertices the polygon is created with. It is preferred for annotation of objects when a rough estimate of each object's location suffices. To allow a pixel-precise annotation, a binary mask is superior. A binary mask is a high-dimensional data point with the same number of rows and columns as the width and height of the annotated instance in pixel/voxel units, made up of 1's and 0's. With this distinction in mind, although a binary mask may seem more expensive to store, there exist encoding methods for tackling this very issue, such as the \textit{Run Length Encoding} (RLE) \citep{tsukiyama1986method} (used in \cite{code2014cocoapi}). However, this is only a good solution for storage, not for real-time manipulations (to refine segmentations). This is why most of the existing annotation platforms prefer polygons. It is worth mentioning that some annotation platforms may present the illusion of creating binary masks (e.g., by including brush-like tools in their toolboxes), but the data may still be converted to polygons via some shape approximation algorithms. This is a lossy conversion, that is, the granularity of the shape is not recoverable.
        
        Two other PoIs which are used in more specialized areas of research are \textit{polylines} and \textit{skeletons}. The former is a path, a sequence of points forming connected segments. They are used when the spine, curvature, or the linear structure of an OoI is of interest (e.g., a spine of a filament, as shown in Fig.~\ref{fig:types_of_PoIs}). The latter is an acyclic graph. It is particularly useful for annotating OoIs with elongated or tubular shapes (e.g., branches of trees, roads, wires, etc.) Both of these PoI annotations provide a simplified and concentrated representation of the main structure, reducing complexity. In bio-imaging, skeletons PoIs are useful for the (so-called) `skeletonization', such as the skeletonization of animals' airway trees \citep{ghanavati2014automatic}. Storing the skeleton of an object is typically more space-efficient than storing the full OoI's binary mask. Creating such PoIs is also often faster and less labor-intensive than annotating the entire OoI instances. This feature makes this PoI particularly useful for tracking and movement analysis as well. Needless to say that it can lose detailed information about the OoIs if they are not coupled with other types of PoIs.

    \subsection{End Goals}\label{subsec:end_goal}
        The end goals of an annotation project are basically the objectives defined for another project which uses the manually annotated data. For example, the BRATS dataset provides the annotations of brain tumors in MRI images to satisfy the end goals of automatic localization and classification of brain tumors from MRI images \citep{menze2015multimodal}. With today's ubiquitous AI applications, the general end goals of almost any image annotation project is arguably to serve machine learning algorithms which are in turn are utilized to automate and speed up the annotation process. As is the case in any project, the team should make sure their efforts are effectively serving the end goals of the project. This is even more true for complex projects such as the manual annotation of scientific images.
        
        We would like to emphasize a crucial point, as we will refer to it in this paper a few times. \textit{Any decision concerning the annotation task should be made in the light of its direct or indirect impacts on the end goal(s) of the project}. Manual annotation of image data, as mentioned above, is typically done to serve machine learning algorithms in classification, localization, and analysis of objects or events, and to build an automated system that carries out this task much faster than and as good as---if not better than---human. However, this common objective has many unique details which separate one project from the other. Therefore, at the early stage of the project timeline, the end goals of the project must be laid out and discussed thoroughly by the team, and then be documented as accurately as possible, and disseminated to all team members, to ensure that any deviation from the end goals is minimized.
        
\begin{figure*}[t]
            \includegraphics[width=0.8\linewidth]{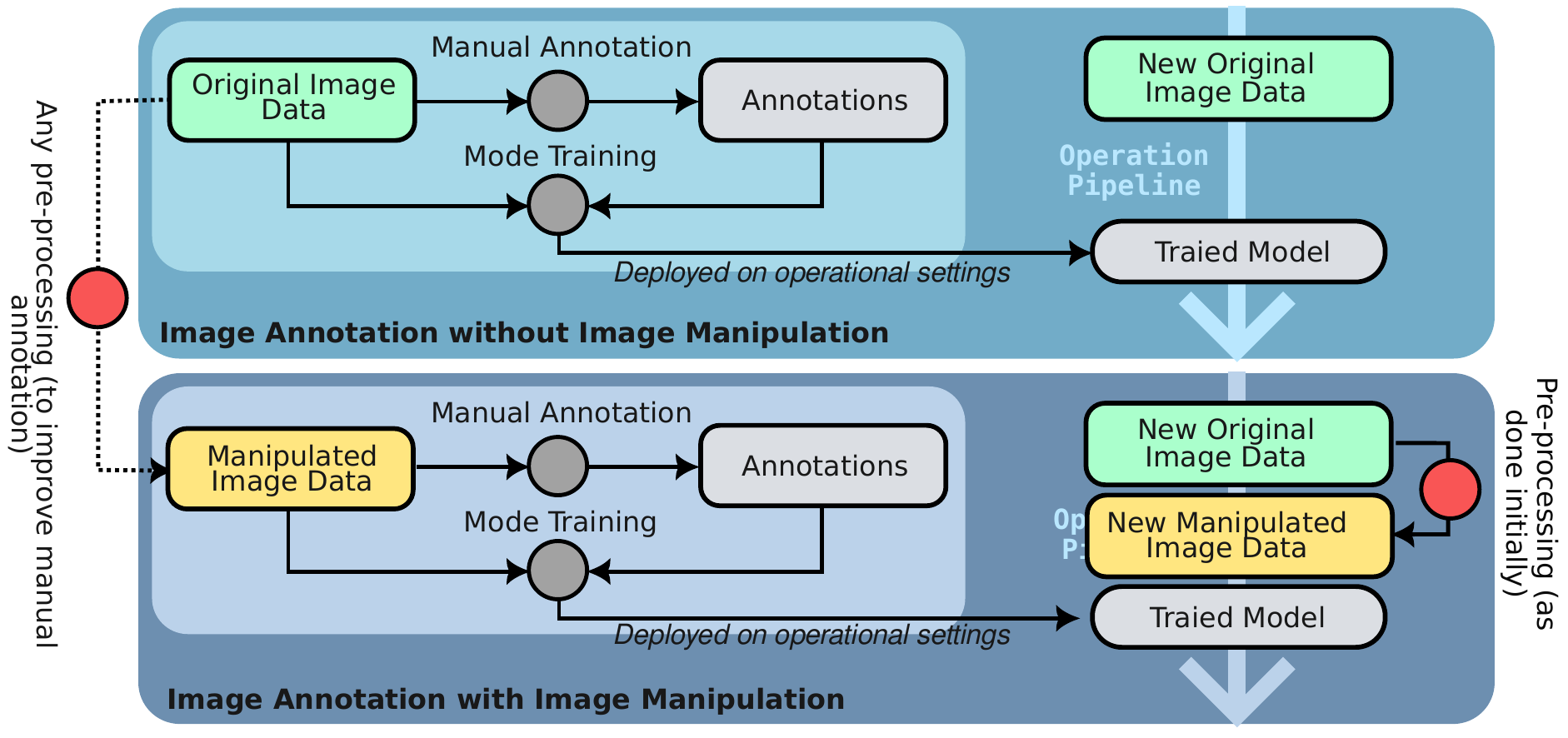}
            \centering
            \caption{The figure shows the differences in operational setting when any image pre-processing is used in order to help the annotators produce more confident annotations. The red disk indicates the pre-processing algorithm and the fact that the exact same process must be incorporated in the Operation Pipeline because it was used in the Manual Annotation process.}
            \label{fig:risk_of_image_manipulation}
        \end{figure*}

        To give an example, any manipulation of the image data to enhance the annotation process must be decided in the context of the end goals. If the team decides to convert their images from a scientific format to a more user-friendly format (e.g., JPG or PNG), they must examine the impact of this decision on the operational object-detection systems. The conversion cost, for instance, might become a computation bottleneck for the online systems which receive continuously generated data in the original format. Overlooking such a critical factor may potentially render the annotated data less valuable because of the discrepancies between the training data (i.e., annotations made on the converted images) and the testing data (i.e., annotations to be made on the original images). Fig.~\ref{fig:risk_of_image_manipulation} shows the differences in the operation pipelines if any pre-processing algorithm (such as the conversion algorithm mentioned in this example) is to be used prior to the manual annotation process. Note that in the top section of the schema, no image manipulation was carried out and therefore, the Operation Pipeline can pass forward the original images. However, in the bottom section, since image data were previously manipulated during the Manual Annotation process, the deployed Trained Model expects to receive images which are identically manipulated in the Operation Pipeline. 

\section{Data to Annotate}\label{sec:data}
    
    A thorough review of the images to be annotated is essential for identifying any of the issues which may cause challenges when the manual annotation phase begins. In this section, we review some of the data-related concerns with respect to the state in which they may occur. 
    
    \subsection{Basic Properties of Images}\label{subsec:basic-properties}
        Images are high-dimensional data with unique properties such as size, resolution, pixel-intensity distribution, and format. The annotation team must include these aspects in the data quality and readiness assessment stage.

        For datasets containing images of different sizes, there might be concerns regarding those which are too small or too large. Very small images may not be appropriate for a high-precision annotation task, and therefore, their exclusion should be discussed. For very large images, on the other hand, a few other factors should be assessed, such as (1) the storage and memory capacity of the annotation platform of choice for handling such images (see Sec.~\ref{subsec:annotation-platforms} for more details regarding the platform preferences), (2) the memory capacity of the devices accessible to the annotators, and (3) (for web-based platforms) the Internet bandwidth accessible to the annotators.

        Image resolution, another important basic property of image data, is sometimes loosely equated with the image size. The term (spatial) \textit{resolution} refers to the number of pixels produced by the CCD array in a camera and is measured by the unit \textit{pixels per inch} (PPI) \citep{acharya2005image}. The size of digital images however, may be manipulated using computer programs while their inherent resolution (i.e., the amount of details in one unit of space) remains unchanged or decreases. The annotation team should be careful about such a resolution variance in their image data, as this may raise some challenges in the annotation phase. Suppose, for example, annotators are expected to capture the jagged patterns of type-A leaves, which differentiate them from type-B leaves; leaves without any jaggedness. But in low-resolution images, such patterns may not be visible for either of those leaf types. Without any proper remedy, the annotators have no choice but to annotate both type-A and type-B leaves as if they belong to type-B OoI. Such bad practices add bias to the annotated data, lowering the upper-bound performance for the machine learning task. The microvascular abnormality detection, blood vessel tracking, and lesion identification are some of the more common areas in medical research, which may be impacted by such practices, pose significant hurdles during the annotation phase, and increase annotation time and error even for seasoned experts.

        In scientific domains, the images are digitized into non-typical image formats. As we discuss in Sec.~\ref{subsec:preprocessing}, non-typical image formats introduce a new set of challenges. The team should anticipate if they can keep the same format across the different stages of the annotation process (e.g., whether the chosen annotation platform supports such formats). If that is not possible or too costly, they should design a conversion algorithm and thoroughly review the ramifications of this conversion (e.g., loss of some textural features and the additional computation time). They should also take into consideration that their targeted audience may not feel comfortable with the decided image format, often due to the loss of some information as a result of the conversion. We discuss these aspects in more detail in Sec.~\ref{subsec:preprocessing}.

    \subsection{Data Availability}\label{subsec:data_availability}
        Prior to the design of a data sampling algorithm, the team should investigate the data sources and the availability of the data. As this may cause unforeseen costs and delay in the delivery of the product, we find it very important to make a clear distinction between three states of data availability: (1) \textit{seemingly available} data; the instances are claimed to be accessible to the team, (2) \textit{evidently available} data; the accessibility of instances is verified by the team, and (3) \textit{truly available} data; the accessibility of the instances is verified by the team to follow the exact expected/documented specifications. Below, we elaborate on the differences between these possibilities.
        
        When the team is aware of a collection of data, this awareness often comes with some expectations which should not be presumed. Suppose that a web interface gives access to the targeted data. The interface may simply stop working for an extended period of time without any announcement to the potential users, or any timely fix. In such a case, seemingly available data are not really accessible by the team.
        The annotation team can avoid such unexpected issues by directly accessing the data, checking the historical accessibility of the interface (up/down time), or even getting in touch with the interface's administration team.
        
        Identifying that a dataset is only evidently available requires some effort. The team must explicitly access the data and ensure that they can indeed see instances they expected to see. What may seem to be a natural expectation should always be taken with a grain of salt. An undocumented (or poorly documented) limit on the number, quality, or format of available images, as well as the team's access to the recent images versus only the historical images, are some of the key factors in rendering a dataset not truly available (only evidently available). Each of such factors may leave the project with a significant setback. 
        
        Ensuring the true availability of data may become complex at times. The only appropriate way to deal with this challenge is by implementing throw-away code snippets and testing the team's each and every need, one by one. It is worth reiterating that, although some of these needs might have already been explicitly addressed in the documentation of the data, it is still highly risky to plan an expensive annotation project solely based on such claims, especially because inaccurate statements in documentations and the ``documentation debt'' are more common than we would like to admit \citep{kruchten2012technical, ampatzoglou2016perception}. As an example of an evidently available dataset which is not necessarily truly available, consider an Application Programming Interface (API) which allows retrieval of astronomical observations made by different observatories. The observations are made at a fixed cadence for years. Now suppose that the API allows users to retrieve images following different sampling strategies by specifying the instruments' names, the sampling time window, and the sampling cadence. While this API may be sufficient for some sampling methods, an undocumented limit on the number of images which can be retrieved by one query request can change the scene. The team may call the API and test the true availability of the data, yet their query may not be just expensive enough to pass the threshold. If a query exceeds the limit (but no proper messages are communicated to the users), the team may not even notice that they are missing some observations. If they do notice the missing observations, challenges like this may have simple fixes. In this case, breaking longer periods of search queries into smaller ones may resolve the problem. However, it is not entirely unlikely that a combination of unforeseen factors makes it impossible, or very expensive, to implement an unbiased sampling method. If a cap on the number of samples to be queried limits the users in their sampling attempt, then the team may find themselves in a situation where they have to recreate the entire database on their local systems (transferring petabytes of image data) to gain full control over the data and achieve a large and statistically unbiased sample.

        The ways that the true availability of data may be compromised vary greatly from one project to another. It is infeasible to catalog all such possibilities generically. It is up to the annotation team to anticipate such possibilities and devise methods for verifying the availability of the data.

    \subsection{Data Sampling Strategy}\label{subsec:data-sampling}

        \begin{figure*}[t]
            \includegraphics[width=0.6\linewidth]{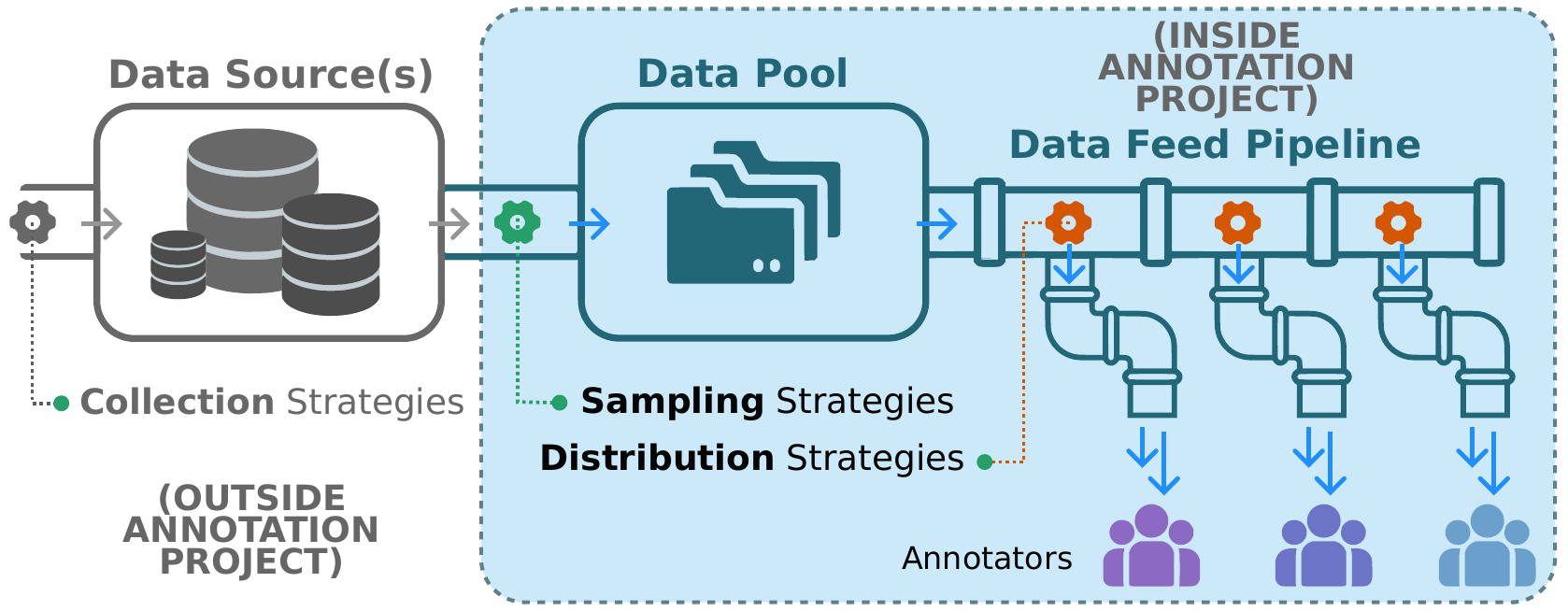}
            \centering
            \caption{The figure illustrates the relation between a \textit{data source}, a \textit{data pool}, and a \textit{data feed pipeline}.}
            \label{fig:data_source_pool_feed_pipeline}
        \end{figure*}
        
        Data sampling, in the context of this paper, refers to the choosing of samples forming a subset of targeted image data to be used for the annotation purpose. We prefer the term `data sampling' over `data collection' as the latter may be confused with the process of \textit{creating} data using sensors, instruments, surveys, or by scraping the web. A manual annotation project does not necessarily need to create a new dataset and is often carried out on existing data. At this point, the annotation team can proceed with developing data sampling strategies.
        

        Let us also draw a distinction between three terms in the context of data annotation: a data source, a data pool, and a data feed pipeline. As illustrated in Figure~\ref{fig:data_source_pool_feed_pipeline}, a \textit{data source} contains the raw data, and it is stored in the infrastructures not necessarily owned or maintained by the annotation team. A data source is usually not designed specifically to serve the annotation team, but to serve a broader audience. Unlike data sources, a \textit{data pool} which is simply a dataset, is prepared by the team (following certain sampling and pre-processing strategies) and is stored in the team's storage infrastructure. Any image pre-processing tasks can be carried out during creation of a data pool. A data pool supplies pre-processed image data (see Sec.~\ref{subsec:preprocessing}) to the data feed pipeline (introduced next), at once, iteratively in batches, or continuously. A \textit{data feed pipeline} in an annotation project is a process which brings annotation-ready image data to the annotators' end point (following certain distribution strategies). There is an important distinction to be made between the objective of the sampling strategies and that of the distribution strategies. The sampling strategies focus on achieving homogeneous, usable, OoI-rich (containing enough OoI instances), and annotation-ready images, whereas the distribution strategies focus primarily on balancing the load between the annotators and enhancing the quality and quantity of the annotations. For balancing the load, the data distribution strategies must also take into account the distribution of OoIs instances and the expected PoIs in each image. Lastly, as we discuss in Sec.~\ref{sec:bias} in more depth, each of these three phases contribute heavily in the introduction and/or mitigation of bias in the annotation project.
        
        This distinction is drawn because it is tempting to equate a data source with a data pool that is prepared for a particular annotation task. For example, NASA's spacecraft, Sentinel-6 Michael Freilich \citep{donlon2021copernicus}, provides a stream of satellite data monitoring the sea level, i.e., it may serve as a data source for an annotation project. The annotation team may decide to treat the existing API to directly pump data to the annotators' end points, disregarding the need for a data pool. However, this imposes a number of limitations. First, the data sampling strategies (now replaced with the API) will be then limited to the queries already incorporated in the API. That is, if the sampling strategy requires integration of an extra piece of information other than those already collected by the Sentinel's instruments (e.g., measures of solar activities \citep{roy2010solar}), the pipelining strategy may fail to meet the sampling rules. Second, the data distribution strategies (also replaced with the API; see Sec.~\ref{subsec:data-distribution-strategy}) may be designed in the absence of a holistic view of the data. That is, the images may be assigned to the annotators without taking into account whether the statistical distribution of OoIs is uniform (see Sec.~\ref{subsec:bias-in-sampling}). Consequently, such a data distribution strategy may initially overload the annotators with OoI instances due to the high prevalence of such instances, and, after a while, create a dry spell of OoI instances. Similarly, it could distribute a few OoIs more often than the others during a period, influencing the annotators with the assumption that some OoIs are more prevalent than others.

        Another important aspect of data sampling, that must be paid attention to, is the number of data sources from which the samples are expected to be retrieved. If the team is invested in only one (homogeneous) data source, the sampling strategies should be able to collect (1) the expected number of images, (2) the expected number of instances of each OoI, and (3) a reasonable percentage of usable images (i.e., those which are not corrupted or of unacceptable sizes, ratios, or resolutions). Further, the sampling strategies must minimize the bias listed in Sec.~\ref{subsec:bias-in-sampling}.
        
        If the team is invested in more than one data source, in addition to the concerns mentioned above, the data heterogeneity should be carefully examined and homogenized. Heterogeneous data (as opposed to homogeneous data) are ``any data with high variability of data types and formats'' \citep{wang2017heterogenous}. For instance, microscopic images captured by different devices---and potentially by different teams under significantly different lab environments---may have different resolutions, lighting, instrument maintenance protocols, etc. Depending on the end goals, the annotation team may need to consider pre-processing of the images in order to unify the pixel intensity distributions across all devices. Such methods are highly domain-specific and must be thoroughly discussed with the domain experts. The development team should also be engaged in such discussions prior to any actions to ensure that the task is not beyond their resources or the project's timeline. To reiterate what was discussed in Sec.~\ref{subsec:end_goal}, the impact of such manipulations on the end goals must be carefully assessed, noting that the annotators will be working on the manipulated images and not the original ones.

        To design the data sampling strategies, the team should lay out a set of explicit, agreed-upon rules describing the desired and undesired properties of the collected data, including those addressing the concerns discussed above. The objective of this task is that the data feed pipeline executes the data distribution strategy (see Sec.~\ref{subsec:data-distribution-strategy}) without any significant sacrifices. Each of such rules may bear its own technical challenges which should be carefully examined. Suppose an annotation team working on satellite images to identify colonies of a species and the colonies' sizes, decides to have a (sampling) rule requiring a temporally uniform sampling of images. The team's justification may be to minimize the impact of seasonal bias on their data, however, due to the migration patterns of the species the rule may yield a data pool with an extremely scarce presence of the targeted species. This imbalance, if not addressed in the distribution strategy (see Sec.~\ref{subsec:data-distribution-strategy}) may cause waste of resources during the manual annotation phase as the annotators would have to spend quite some time finding any instances of the scarce OoI to annotate. Such complications and their consequences should be spotted as best and as early as the team can. Although miscalculations of future needs are part of any real-world project, a hasty approach at this stage may snowball the cost at later stages. 
        
    \subsection{Pre-processing of Images}\label{subsec:preprocessing}
        Although, initially it may seem that the targeted image data are annotation-ready, they might not yet be. There are a number of different considerations that the team should carefully examine before they can move on to the next step.

        Perhaps the simplest item to check concerns the image format of the images in the target data. Digitized astronomical observations are often stored in FITS\footnote{FITS stands for Flexible Image Transport System.} format. MRI devices use different (proprietary) formats depending on their manufacturers, such as PAR/REC or IMA files by Philips and Siemens scanners, respectively. There are numerous application-specific file formats used for capturing 2D or 3D images of OoI. The annotation team should first investigate whether the annotation platform of choice can process such non-typical image formats. Consequently, if the team eventually decides to convert the images to a platform-friendly format, the conversion algorithm must be scrutinized by domain experts to ensure that the data loss does not impact the annotation. This is in addition to the conversion cost we mentioned earlier in Sec.~\ref{subsec:end_goal}. Revisiting the point we made in that section, it should be noted that the annotation setting should be similar to that in the operational setting. One should not assume that an annotator would create identical PoIs on images regardless of the format they are presented in. For example, when the annotators are asked to rely on subtle textural information to classify objects, such pieces of information may systematically change or be lost during a (lossy) conversion process. 

        Even if the annotation platform of choice is capable of processing the non-typical image format that the team needs to annotate, one should be reminded that these formats are non-typical for a reason. The inclusion of extra information about the imaging process (time of capture, exposure time, instrument name or model, etc.) is one of such reasons. Indeed, another important consideration is that the pixel intensity range in certain image formats deviates from the standard 0 to 255 range found in typical 8-bit quantized image formats. That range is determined by the sensitivity of the CCD sensor (in CCD cameras) and represents how many different intensity levels each pixel can represent. Therefore, a linear transformation may not always result in the best visual presentation of the OoIs (see \cite{Ahmadzadeh2019curated} for a detailed discussion on using JP2 format instead of FITS for the classification of solar events). The way the annotation platform is programmed to show these images to the annotator may not completely agree with what the domain experts desire, especially because such agreements change from one task to another. Therefore, the possibility of incorporating an in-house conversion stage prior to using the annotation platform should be considered.

        Another factor to keep in mind when planning for the pre-processing stage is that scientific image datasets are generally expected to follow certain unifying rules. Whenever possible, it is expected that corresponding image features are spatially aligned. For example, in MRI images of the brain, image transformations are used to correct for spatial discrepancies (such as rotation, translation, etc.) caused during the scan. This process is generally called ``image registration'' \citep{hajnal2001medical, moigne2011image}. In satellite imagery, image registration is also used to correct for area-based or feature-based, geometric distortions \citep{eastmansurvey2011, dave2015survey}. Depending on the readiness of the targeted data, such pre-processing steps may or may not have already been carried out. It is critical to assess (1) whether such steps impact the annotation process in any way, and (2) whether the annotated data can be transformed later along with the images when (and if) they need to undergo any correction/manipulation for future registrations towards satisfying the end goals. 
               
        Verifying that the rules defined during data sampling (see Sec.~\ref{subsec:data-sampling}) were actually enforced is another critical factor that can be carried out in the pre-processing stage. This fits in this stage because some images may need to be removed due to some issues. For example, the team can verify whether there is at least one OoI instance present in each image. If so, such images can be removed or replaced with others. Doing so avoids confusing the annotators, and more importantly, it guarantees a relatively fair distribution of workload among them.
    
        Pre-processing of images is a stage where the team can benefit the most from the close engagement of their domain experts. The numerous nuances that domain experts can investigate in the data are highly subject-specific and beyond the scope of this domain-independent paper. As a simple example, consider chest X-ray image data. The angle and distance of the patients from the X-ray source, regardless of the body position at which the images were taken, may influence the chance of (mis-)diagnosing Cardiomegaly---a medical condition in which the heart becomes enlarged (see the pre-processing discussion in \cite{bustos2020padchest}). Such natural inconsistencies in data are very common, and depending on the application, they may or may not need to be addressed in the pre-processing stage. Without proper engagement from domain experts in the annotation project, critical pre-processing practices may be overlooked, potentially incurring significant costs for the project.
        
    \subsection{Data Distribution Strategy}\label{subsec:data-distribution-strategy}
        When images are ready for manual annotation, they should then be distributed among the annotators through the data feed pipeline. This phase requires a careful design of data distribution strategies to enhance the efficiency and reliability of the annotation pipeline. Below we discuss some of the key aspects needed in such a design.
        
        \textbf{Maximizing the manual annotation efficiency} - A bad data distribution strategy can indeed slow down the manual annotation process. The simplest design flaw that can cause this is where the presence of OoI instances in images is not taken into account. This may result in a distribution which contains a very scarce presence of OoIs, at least in a subset of the images. The long browsing time spent by the annotators for finding OoI instances to annotate is wasteful and increases the total annotation cost. Another design flaw (with a somewhat similar nature) takes place when the number of OoI instances assigned to each annotator is not proportional to the number of images assigned to them. This imbalance leaves some annotators with a significantly heavier workload than others. Furthermore, given that some OoIs may require more time and effort than others, this could also unfairly assign a larger portion of more difficult OoI instances to some annotators, leaving the easier instances to the rest. For example, any OoI instance which requires that the annotator consults the provided supplementary sources (see Sec.~\ref{subsec:supplemental}) takes significantly longer annotation time. Since the compensation of the annotators is often measured based on the number of images or OoI instances annotated, this can create an uncontrolled variance in the labor value. The team should be aware of such discrepancies and find appropriate solutions or workarounds for maintaining an efficient and sustainable annotation pipeline.

        \textbf{Minimizing the embedded biases} - In Sec.~\ref{sec:bias}, we discuss the different ways bias can permeate the annotation project, as well as costly consequences. More specifically, in Sec.~\ref{subsec:bias-in-data-distribution} we discussed bias in data distribution. Therefore, we refrain from repetition and refer the reader to those sections.
        
        \textbf{Laying out an aggregation strategy} - The motivation behind any data distribution strategy is not only to divide the work among the annotators but also to aggregate multiple views into one and build more reliable annotations. Depending on the specifics in each project various aggregation strategies may be adopted. We would like to review three general strategies:

        \begin{itemize}
            \item \textit{Best Overrides Rest.} In this strategy, multiple annotations are collected per PoI of each object and the final annotation is chosen or inferred---according to a set of pre-defined aggregation rules---to replace the others. For the classification task (i.e., no PoI), this is often referred to as the majority vote \citep{deng2009imagenet}. For PoIs, the aggregation rules should be specifically defined after consulting with the domain experts. For the coordinate PoI, the best coordinate per OoI instance may be defined as the center of all the manually identified coordinates for that instance. For the bounding-box PoI, a few simple-to-generate options are to take (1) the intersection, (2) the largest, (3) the smallest, or (4) the bounding-box spanning over all boxes corresponding to each OoI instance. For segmentation PoI, although similar options work, they may not always be desirable. Note that the purpose of collecting segmentation PoIs from scientific OoIs might be to identify them based on their unique shape structures (e.g., to identify different plants based on their leaves' unique shape structures). In such cases, the choices we listed for the bounding-box PoIs may result in loss of data or introduction of noise. For instance, the intersection/union between two shapes, where one is nested inside the other, reveals nothing about the characteristics of the outer/inner shape. Some alternatives however exist, to mention a few: (1) the number of vertices (if segmentations are stored as polygons) can serve as a proxy for the granularity of segmentation that annotators have incorporated, (2) a finer or a coarser structure, depending on the task, may also be a good choice, (3) the fractal dimension of segmentations \citep{theiler1990estimating, mandelbrot1982fractal, barnsley1988science} which is a geometrical property for shapes, can also capture segmentations' granularity for cases where they are stored as binary masks.

            The main advantage of the Best-Overrides-Rest strategy is that it allows having a quantitative quality control system. As multiple annotators work over each instance of OoI, the overall agreement can serve as a proxy indicating the confidence of the annotations. If need be, the team may intervene and conduct retraining; review the common issues in the current annotations and the best practices. Another advantage of this strategy is that it does not require a review process (where domain experts serve as reviewers); this is generally easier and cheaper to implement than the Redo-Until-Accepted strategy that comes next.

            On the downside, this strategy lacks a qualitative quality control aspect which is instrumental for producing a reliable annotation of complex OoIs. When it comes to scientific imagery, devising quantitative measures for accurately and systematically capturing whether an instance is annotated as expected or not is not a trivial task. Therefore, the final quality of the annotated data is, at best, as good as the quality of the best annotators' work in each group. Given that this strategy does not allow for the annotations to be returned to the annotators with some constructive feedback, the quality is expected to regress to `acceptable' instead of `best'.

            \item \textit{Redo Until Accepted.} In this strategy, a set of instances is assigned to one, and only one, annotator and each instance will be returned to the annotator if it does not satisfy the requirements. There are situations in which the Best-Overrides-Rest strategy may not be feasible. This is particularly the case when the subjective notion of `correct` is hard to be defined, i.e., multiple (and possibly, infinitely many) instances of PoIs can be equally satisfactory at the same time. For instance, in the annotation of streets in aerial imagery, the area where a street ends and the sidewalk begins is often below the image resolution and therefore, appears as a gray strip. What portion of this gray strip belongs to the street's segmentation may be considered a judgment call. So, there are many correct segmentations and yet, there are many incorrect ones as well (e.g., an annotator may include only one side of a street in their segmentation). The Redo-Until-Accepted strategy may be more appropriate in such cases.

            The main advantage of using this strategy is the direct supervision that it brings to the table. Instead of relying on the majority vote and aggregation strategies, the annotations must satisfy the reviewers, i.e., the domain experts. This qualitative aspect is what makes the final product much more reliable, compared to the Best-Overrides-Rest strategy which relies on the quantitative agreements. Quantitative measures may fall short when it comes to the annotation of complex OoIs (see the discussion in \citep{ahmadzadeh2021multiscale} about the inadequacy of some popular similarity measures). 
            
            It is worth mentioning that the Redo-Until-Accepted strategy requires only one annotator to work on each instance. Therefore, it may seem a faster approach compared to the Best-Overrides-Rest strategy. However, the process's speed depends on how rigorous and strict the review process is intended or needed to be. In reality, at least initially when the annotators are not yet well versed in the annotation task, this strategy is certainly slower than the Best-Overrides-Rest strategy. To speed it up, the team can (1) provide the annotators with feedback on the returned instances and (2) keep the same (trained and experienced) annotators on the team for as long as possible, even if a higher compensation rate is needed.

            Another valuable outcome of this strategy is the enhanced training quality. The annotators become extremely good at annotation of OoIs after a few rounds of seeing the feedback provided to them by the domain experts. Such annotators are instrumental to the quality of the final annotation product. It is at the interest of the project manager(s) to keep them in the team even at higher costs than initially negotiated. Note that in some areas, such annotators may quickly become good enough to serve as the reviewers, reducing the workload of the domain experts, and consequently, reducing the overall annotation cost.

            \item \textit{Mixed.} A mixed method can be utilized too. In this strategy, although multiple annotators are assigned to work on one batch of images (i.e., Best Overrides Rest), each annotator's work is reviewed separately (Redo Until Accepted). The annotations are returned to the annotators (rejected) until the requirements are satisfied. At the end, during the post-processing phase, some aggregation rules can be entertained to improve the quality of the annotations. For example, the quality of the segmentation PoIs can be scrutinized during the review process, whereas the assigned classes can be automatically corrected by a majority vote rule. 

            The Mixed strategy has all the benefits of the Best-Overrides-Rest and the Redo-Until-Accepted strategies. Note that, although in the mixed strategy each instance is annotated by multiple annotators, this is not solely for the purpose of cross comparison. Because each instance is separately reviewed and accepted, they all are valuable data points. For machine learning purposes, although multiple annotations correspond to the same instances of OoI, models can still learn from the variance in the annotation data., Such variances improve the robustness of the models.
        \end{itemize}
        
    \subsection{Supplemental Sources of Information}\label{subsec:supplemental}
        Image annotation in scientific domains is often more complex than identifying common objects with conspicuous shapes. The distinction between a malignant tissue and a benign one might be only visible to trained eyes; where to draw the line between an `unhealthy' leaf and a `healthy' one requires training. In some cases, domain experts themselves may need to look at some other pieces of evidence to be confident in their judgement. If such supplementary pieces of information can potentially help the annotators, it might be worth the extra effort to incorporate them into the annotation ecosystem to assist the annotators (see Sec.~\ref{sec:bias} for potentially introducing biases into the annotation workflow).
        
        As an example, in one of NASA's Citizen Science projects named Solar Active Region Spotter \citep{citizenscience2022ARSpotter}, the annotators are provided with two sources of data for each observation: the magnetograms (maps of the magnetic field of the Sun's surface) and the EUV composite (showing solar plasma). In their training web page, the EUV data are introduced as ``supporting data only''. For the annotation of images, another common practice is to provide short videos as a supplementary source of information. This is especially helpful if the OoI instances are heavily surrounded by background noise (e.g., very common in telescopic, microscopic, satellite, and aerial imagery). Watching a sequence of images would help the annotators immensely in separating the background noise from the details of instances of OoI. Solar Jet Hunter (also a NASA's Citizen Science project) takes advantage of successions of images as a supplementary source of information \citep{citizenscience2022JetHunter}. Detecting a jet---a tiny bright line indicating hot material moving away from the Sun---requires looking at the evolution of the solar activities over time. Expecting the annotators to rely on still images would significantly impact the reliability of the annotations.

        The team should use caution, however, when providing such supplemental sources of information to the annotators. Adding extra pieces of information can potentially lead the annotators to adopt theories which are wrong, or under investigation by this very project, introducing bias to the annotation process. For example, suppose that based on a theory, that is not yet tested on sufficiently large datasets, some solar activities are more likely to take place in the northern hemisphere of the Sun than the southern hemisphere. If such a piece of information is exposed to the annotators, the annotators may label solar activities mostly based on their location, instead of their other properties, such as shape or texture. In this scenario, the annotated dataset will lose its value in general, but significantly so for being used for testing that very theory.

        Another important factor that needs to be assessed before adding any supplemental sources is the added time to the annotation pipeline. For the manual annotation of large datasets, extra seconds spent on an object adds up very quickly. Expecting the annotator to consult another source of information to make better decisions adds seconds or minutes to the annotation of each object. Therefore, such a requirement must be kept at a minimum, and be assisted in such a way that it takes the least amount of time for the annotators to complete it. The team should also consider the time needed to develop a platform (if it does not already exist) to provide the supplemental information to the annotators in an efficient way. Development of a user-friendly and bug-free platform is a project on its own, and requires resources and expertise in the team.
        
\section{Annotation Ecosystem}\label{sec:ecosystem}
    The annotation ecosystem encompasses all the components (software products and services) necessary to support and streamline the annotation workflow, ensuring that the team efficiently satisfying the measures of success defined in Sec.~\ref{sec:success}. It is not imperative that all the needed components are fully integrable, as long as the necessary components are compatible, i.e., can easily communicate with each other and pass data and metadata. To build an effective annotation ecosystem the goal should be maximizing quality, consistency, and scalability.
    
    \subsection{Annotation Methods and Platforms}\label{subsec:annotation-platforms}
        To speed up the laborious task of manual annotation, the idea of interactive segmentation was entertained two decades ago \citep{boykov2001interactive, rother2004grabcut}. As a result, annotation platforms are now taking advantage of algorithms to automatically generate first-draft annotations and let human annotators make improvements only as needed. Such algorithms are based on either the classical computer vision approaches such as clustering and thresholding, or the pre-trained AI models (e.g., Meta's Segment Anything Model \citep{kirillov2023segment}). Nonetheless, despite their tremendous success in speeding up the annotation process, their contribution to the annotation of scientific imagery is somewhat limited. This is simply because generic segmentation models do not yet perform meaningfully well on highly specialized tasks in which a pixel-precise annotation of scientific objects/events is expected (e.g., see the conversation in \cite{ahmadzadeh2021multiscale}). Such segmentations are often achieved following a complex set of criteria laid out by domain experts. This is particularly true in the context of highly noisy images where generic algorithms are not prepared for. 
        
        Therefore, a need for feature-rich, user-friendly software products and services have been recognized. As a result, the few existing free and open-source products (such as LabelMe \citep{russell2008labelme,labelme}, VoTT \citep{vott}, and LabelImg \citep{labelimg}) gave their place to paid services which naturally brought an array of advanced features to the users' finger tip and made it possible for millions of images to be annotated and reviewed by human annotators, with the possibility of using AI back-ends. Table~\ref{tab:annotation-tools} lists a few of those tools for the interested readers to compare their functionalities. What is exciting about these platforms, as far as this paper is concerned, is not the AI assisted annotation, but the rich annotation toolkit and the customizable workflows. The annotation team should spend enough time on deciding what platform to choose. A few of the key questions to ask the team are: (1) whether the platform processes their specific image format(s), (2) whether it can generate the expected PoIs, (3) whether it is compatible with their annotation ecosystem (e.g., their cloud data storage), (4) whether it provides the support for training and overseeing the human annotators, and (5) whether it is affordable, taking into account all the associated costs.  

        \begin{sidewaystable*}[htbp]
            \centering\footnotesize
            \begin{tabularx}{\textwidth}{>{\raggedright}p{1.0cm}>{\raggedright}p{1.4cm}>{\raggedright}p{1.4cm}>{\raggedright}p{2cm}>{\RaggedRight}p{1.2cm}*{4}{>{\RaggedRight}X}}                
            
            \toprule
                \textbf{Platform} & \textbf{Deployment Mode} & \textbf{Annotation\newline Mode} & \textbf{Available PoIs} &  \textbf{Import/ Export} & \textbf{Capabilities} & \textbf{Acceptable\newline Data Types}\\
                \midrule
                

                \textbf{VGG Image \newline Annotator \citep{vggImageAnnotator}} & local/ online & manual & point, bbox, circle, ellipse, polyline, polygon & GUI & good for small projects; does not have many annotation methods; \textbf{free \& open source} & images (jpeg, png, bmp, gif, svg, webp), video (mp4, mov)\\
                \addlinespace[0.2cm]
                
                \textbf{Label Studio \citep{labelstudio}} & local (online in dev. phase) & manual & bbox, keypoints, polygon, segmentation (brush) & GUI/API & cloud service integration for storage \& ML back-end; \textbf{free \& open source}, rich documentation & images (jpg, png, svg, webp, bmp, gif), video (mp4, webm)\\
                \addlinespace[0.2cm]                

                \textbf{CVAT \citep{cvat}} & online/ local & manual \& AI assisted & point, bbox, polygons, skeletons, cuboids, trajectories &  GUI/API/\newline CLI & easy GUI, collaboration features, task assignment, progress tracking, \textbf{Open source}, rich documentation & images (jpeg, png, bmp, gif, tiff), video (mp4, mov)\\
                \addlinespace[0.2cm] 
                
                \textbf{LabelBox \citep{labelbox}} & online & manual \& AI assisted & point, bbox, polygon, cuboid, segmentation brush), polyline & GUI/API/\newline CLI & easy GUI, collaboration features, performance dashboard, annotation automation, dynamic workflows, cloud services integration & images (jpeg, png, bmp, pdf), video (mp4, mov) \\
                \addlinespace[0.2cm]                
                
                \textbf{V7 \citep{V7}} & online & manual \& AI assisted & bbox, line, ellipse, polygon, keypoints, skeleton, segmentation (brush) & GUI/API/\newline CLI  & easy GUI, collaboration features, performance dashboard, annotation automation, rich documentation, cloud services integration & images (jpeg, bmp, png, pdf, tif, tiff, webp), videos (avi, mkv, mov, mp4)\\
                \addlinespace[0.2cm] 
            
                \textbf{Scale AI \citep{scaleAI}} & online & manual \& AI assisted & polygon, segmentation (brush) & GUI/API/\newline CLI & cloud storage integration, rich documentation & images (Not specified), video (Not specified)\\
                \addlinespace[0.2cm]
        
                \textbf{Super-  \newline Annotate \citep{superAnnotate}} & online & AI assisted & point, bbox, ellipse, polygon, cuboid &  GUI/API/\newline CLI & easy GUI, collaboration features, Performance dashboard, annotation automation, task assignment, progress tracking, cloud services integration & images (jpeg, bmp, png, pdf, tif, tiff, webp), videos (avi, mkv, mov, mp4)\\
                \addlinespace[0.2cm]
        
                \textbf{Dataloop \citep{dataloop}} & online & AI assisted & Point, polyline, bbox, ellipse, polygon, segmentation (brush), keypoints, cuboid & GUI/API\newline  & easy GUI, collaboration features, performance dashboard, annotation automation, rich documentation, cloud services integration & images (unspecified), video (unspecified)\\
                \addlinespace[0.2cm]
                
            \end{tabularx}
            
            \caption[]{A list of some annotation platforms. It is worth mentioning that (1) this list is by no means a comprehensive list, (2) this paper is not sponsored by any of these products and the authors do not wish to favor any product(s), and (3) the specified details are subject to change due to the rapid development of these platforms.}
            \label{tab:annotation-tools}
        \end{sidewaystable*}

    \subsection{Use and Compatibility of APIs}\label{subsec:apis}
        A dynamic annotation project enjoys the flexibility of making changes in every phase of the project, from pre-processing to data sampling and distribution strategies, and even in the recruitment strategies. To achieve such a dynamic project, the team can take advantage of the Application Programming Interfaces (APIs) of the tools and products they incorporate in their annotation ecosystem. APIs make it possible to automate different components of the project reducing repetition and human error. An example of such repetitive tasks that can be automated is the routine retrieval and analysis of the annotated data. When automated, the team can easily identify systematic errors without spending time on one-by-one review of the items. 
        
        The typical APIs that the annotation team may need access to (if possible) are (1) an API for assisting the data collection process (e.g., a REST API that allows querying batches of samples from a data source), (2) an API for importing images into the annotation platform and exporting the annotations from the platform (see Table~\ref{tab:annotation-tools}), and (3) an API for feeding the final annotation product to end users (e.g., a web app that allows search and visualization of the annotations; a data loader package for feeding machine learning algorithms for training). The annotation team should think ahead regarding the availability of such APIs (and potentially other APIs that might be needed). 

    \subsection{Use of Project Management Products}\label{subsec:project-management-products}
        The utilization of a few project-management products is instrumental for the success of any large project, and the annotation project is no exception. Without getting into the details of good practices in project management, we strongly encourage the team to consider taking advantage of at least a few essential tools. We will explain where each of those tools help enhance the synergy and productivity of the team and the quality of the final products.

        The first tool in our list is a Git-based source code repository hosting service (e.g., GitHub, SourceForge, Bitbucket and GitLab). It allows the developers of the team to collaboratively develop the software products the team needs, such as the data collection and data sampling software products. If used within a powerful ecosystem (see the next items), it can bring invaluable organization and security to the project through version control.

        A project tracking software (e.g., Jira, Trello, Asana, etc.) can help the entire team to monitor the progress of the project, and create tickets for issues and bugs they encounter (or even potential points of concern). It also makes it possible to retrieve the solutions to specific issues resolved in the past. If compatible with the team's repository hosting service, issues (known as `tickets') and resolutions (known as `commits') can be linked for review by the lead developers.

        A knowledge base system (e.g., Confluence, Notion, YNAW, etc.) where every team can document their work. A well-maintained knowledge base system makes it possible to document (1) set-ups and configurations for different systems and software products, (2) how each challenge is resolved and why, (3) the points that the team should revisit in the future, and (4) the external sources of information (papers, blog posts, tutorials, etc.) which have been used for different purposes. It is worth mentioning that a rich documentation of the project's progress makes the on-boarding of the new team members significantly easier and more efficient. Moreover, the content of this knowledge base system can help immensely the release of the final product as it can facilitate documentation generation process for end users (in the form of an online document or a peer-reviewed paper). Similar to the project tracking software, a compatible tool can be used to dynamically link documentations with the corresponding tickets and code snippets.
        
        A communication platform (e.g., Slack, Discord, etc.) in which team members can quickly interact with the right person and ask their specific questions is another tool. Specific tags can be devised for each user indicating their expertise (e.g. John Smith [Annotation Reviewer], Jane Doe [Data Engineer]). This platform connects the developers to the domain experts so that small decisions can be made instantly. Moreover, when the manual annotation phase starts, it connects the new annotators to the experienced ones and the domain experts who take on the training role of the annotators. In public channels of such platforms, answers to each annotator's questions can serve all the annotators, quickly becoming an extremely valuable starting point for the new annotators. This platform can also be chosen compatible with the other management products, however, this compatibility may not be essential.

\section{Recruitment and Training to Conduct}\label{sec:training}
    While manual annotation of objects is often considered a purely mechanical job, one cannot survive its tediousness without smart management and productivity solutions. Therefore, what prevents the annotation team from falling into the high-employee turnover and a churn-and-burn culture is some investment in the recruitment and training phases, aiming to hire the most suitable candidates.    
    
    \subsection{Starting from Recruitment}
        Let us start with the job ad as it can directly impact the efficiency of the annotation task. First, the job description for annotation must clearly spell out what the job entails (and what it does not). In an academic setting where students are considered for such a job, it might be tempting to present the job as a ``research opportunity''. It should not be difficult, however, to agree that the task will quickly leave the newly recruited annotators with disappointment. Another misleading factor in the job ad might be the monetary compensation. The team may have an unrealistic expectation as to how fast one can annotate a batch of images and therefore list a superficial compensation rate in the ad. For example, the team may not take into account the difference between the annotations made by an annotator and those which are \textit{accepted} by the reviewers. It is best that the estimates are empirically driven from the actual trials carried out by a few candidate annotators. In both examples, the productivity of the recruited annotators are expected to drop very quickly.

        To facilitate the recruitment process, the team may decide to rely on 3rd party companies (e.g., Amazon's Mechanical Turk) which connect human annotators to annotation projects. We would like to bring to the attention of the reader the growing awareness regarding some potentially exploitative aspects of such practices (see \cite{Hao_2023,Velasco_2024} for example). It is up to the annotation team whether they are willing to follow similar practices, and whether their funding sources allow such practices. Also, note that the human annotators on those platforms have a choice between different projects, and are naturally drawn toward the less demanding ones (e.g., image classification), or the higher paying ones. Therefore, the team must have a good understanding of what a competitive rate is for attracting the more-committed annotators to their more-demanding job. Moreover, the lack of a direct communication channel between the team and the annotators (which is intrinsic to those platforms) may make it unrealistic to get acceptable annotations regardless of the other factors if the task is difficult and requires highly trained eyes.
        

        One of the most appealing aspects of the annotation job is its remote and asynchronous nature. An honest job ad can emphasize this aspect and therefore welcome specific groups, e.g., people with disabilities who may look for a side hustle. Doing so not only contributes to a fairer distribution of opportunities, but also increase the chance of recruiting a more committed group of people due to lack of enough alternative side hustles. If the annotation project is an academic effort, the ad can be directly sent to the university's accessibility \& disability services office, for a direct outreach.

        Since the annotation job is naturally an hourly-paid job, the normal recruitment process may not apply. For example, the interview process may have to be simplified into a short e-form. Therefore, some creative methods need to be devised to reduce the false-positive hiring rate. A filtering mechanism may be designed as follows: the team assigns small test-batches of images (e.g., 5 to 10) to the applicants for annotation before training them (relying only on a short tutorial), and accept their applications only when the test-batch is fully annotated with a somewhat acceptable quality. While a small batch incurs minimal costs for the team, it provides both the applicant and the team with a clear understanding of the applicant's suitability for the task. Many applicants may quit the task after only a few attempts, which significantly reduces the resources needed for further training of the annotators and giving them access to different components of the annotation ecosystem.

        Lastly, the team should be prepared for short-term contributions since usually only a small portion of the annotators remain on the project till the end. The recruited annotators may leave the team after only a few hours of contribution. To maintain some degree of productivity, the team should set milestones (e.g., 50 images fully annotated and accepted) for any contribution to be compensated for. This speaks to the importance of having filtering mechanisms such as the test-batch strategy mentioned before. However, it is critical to note that after reaching the first milestone, the annotator has already developed valuable skills and it is inefficient to lose them. Therefore, if the number of annotators leaving the team after one milestone is disproportionately high, the team should rethink their recruitment strategy. More often than not, in such situations their estimate of the job's difficulty and therefore, the compensation rate is unrealistic and should be updated. That said, unpredictability of the length of contribution is a natural part of hourly-paid jobs. Therefore, the team must have an agile strategy for a quick replacement of the annotators who quit, become unresponsive, or are hopelessly inefficient. This is arguably one of the most challenging aspects of any large annotation project.

    \subsection{Training and Training Material}\label{subsec:training}

        Easy-to-follow, comprehensive, and accessible training materials are key to enhancing the quality of annotation and reducing the cost of recruitment and re-training the annotators. Generally, the training resources can be divided into three groups: the (synchronous) training sessions, the online resources, and the communication platforms.

        Each annotator officially starts their annotation job after participating in a training session. A training session may be held after a group of annotators are recruited (and granted access to the necessary tools). Due to the nature of this job, it generally makes sense that such a session is held online, although, this decision may depend on specific settings. Regarding the content of the training session, two key factors stand out. First, the training session should be about training the annotators. It is easy to fill the training session with unnecessarily specialized details about the science behind the OoIs and the scientific objectives of the project. Although, those subjects must be mentioned during the training session so that the annotators understand the value and contribution of their work, the purpose of the training session should not be lost to the trainer. Therefore, the most productive training sessions are the ones which are hands-on and let (at least a few of) the participants annotate a few instances of OoIs during the session. Second, and perhaps counterintuitive, the trainer does not need to be a domain expert. What should determine the most qualified trainer is (1) their teaching skill, and (2) the amount of experience they have with the manual annotation of OoIs---after all, the annotation task is what the participants are to be trained for, most likely without a deep understanding of the domain. At the end of this session, the participants should feel comfortable about how to use the annotation platform, how to access all the supplementary materials, and, perhaps most importantly, who to reach out to if they have questions.
        
        A dedicated website containing training material can be extremely useful for guiding the annotators towards delivering quality and consistency (this may be reflected as a high acceptance rate in the redo-until-accepted aggregation method; see Sec.~\ref{subsec:data-distribution-strategy}). Having such a website available, the annotators can refer to reliable guidelines when they face any difficulties. To give a practical example, the team can provide a catalog of `good' and `bad' annotations, with examples listing typical mistakes---it is difficult to exaggerate the training value of presenting the typical mistakes. Moreover, short screen-recordings of how experienced annotators approach different OoIs can flatten the learning cave significantly for the newbies. Tips on the efficient use of the annotation platform (e.g., keyboard shortcuts of the toolbox) can also directly increase the workflow and productivity, and therefore decrease the average annotation time. This website can also be used for the recognition of the annotators' contribution to make them feel valued and appreciated, which can enhance their dedication and productivity. It is worth noting that such a platform must be easy to maintain, as well as easy to interact with. Ideal platforms for this purpose are those which do not require any development and deployment skills (only drag-and-drop), such as Google Sites, WordPress, Wix, Squarespace. Otherwise, significant resources should be allocated to ensure timely updates, security, and technical support, just to name a few.

        What nicely complements the training website is a communication platform, discussed earlier in Sec.~\ref{subsec:project-management-products}. While the website contains static content, the communication platform permits dynamic training. It serves two essential purposes: (1) all team members can communicate with each other and address questions and identify the less visible issues, and (2) it makes the annotators (who may be contributing to the project entirely remotely) have a sense of belonging to a real project, with real people, and real milestones.
    
        \begin{figure*}[t]
            \includegraphics[width=0.7\linewidth]{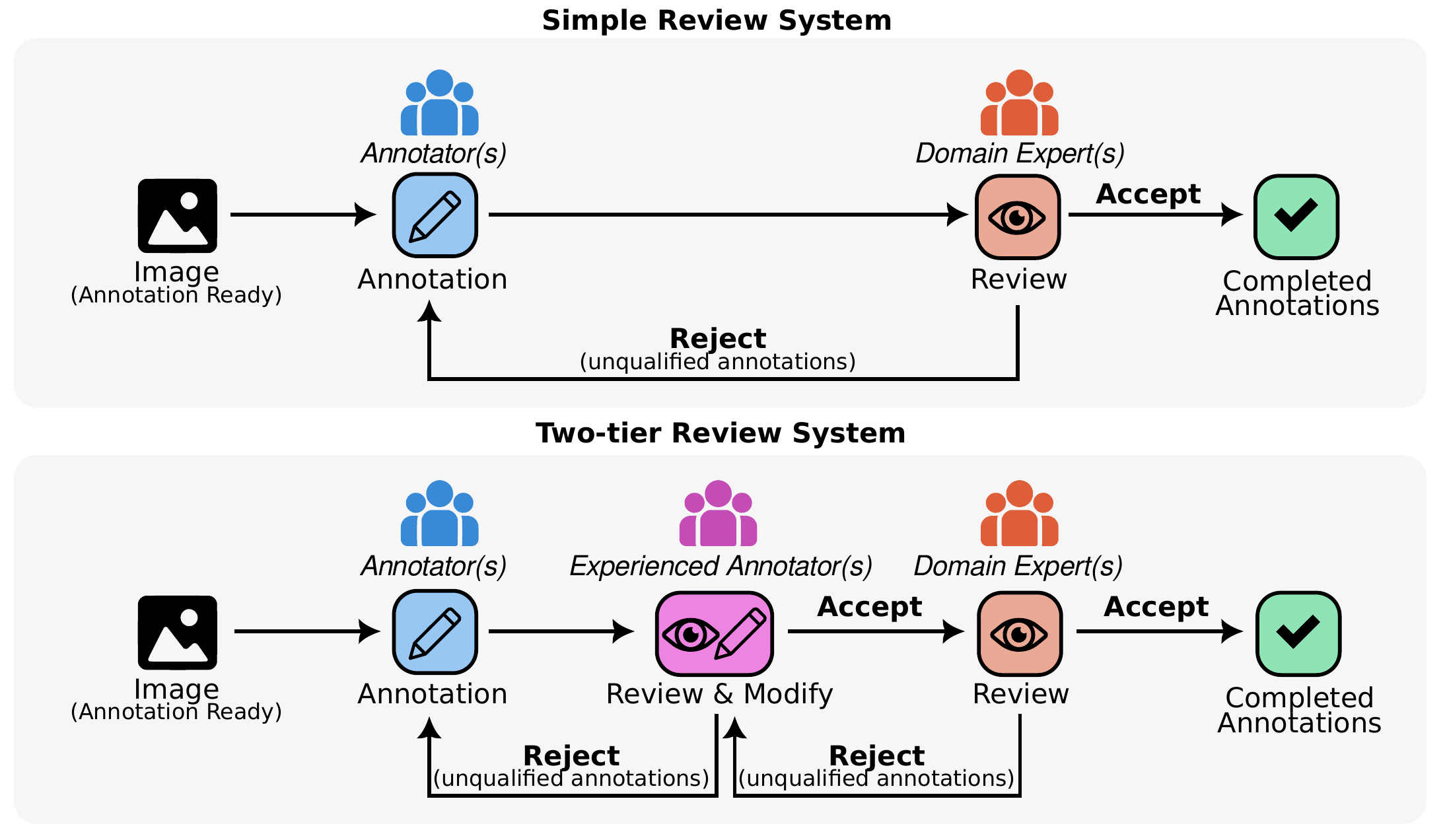}
            \centering
            \caption{The figure shows the schematic of two review systems for manual annotation. On the top, the Simple Review System is shown in which each annotated image is reviewed by domain experts. On the bottom, the Two-tier Review System is illustrated in which each annotated image is first reviewed (and modified if needed) by experienced reviewers, and if qualified, it is passed for review by domain experts.}
            \label{fig:review_workflow}
        \end{figure*}

\section{Bias to Minimize}\label{sec:bias}
    Bias permeates the research process in different ways. It is the team's responsibility (particularly, the data scientists' and domain experts'; see Sec.~\ref{sec:roles}) to identify the exact ways their annotation project might be tainted with bias. Potential biases should be mitigated to the extent possible, and when it is beyond the team's control, they should be documented so that end users are made aware of their possible impacts on their specific applications. In this section, we review some of the typical sources of bias in the annotation project, although, each project's vulnerability to bias is unique and may not be listed in a domain-agnostic paper such as this one.

    \subsection{Bias in Data and Data Collection}\label{subsec:bias-in-data}
        Bias in an annotation project may stem either from neglecting certain intrinsic attributes of the data, or from adopting bad practices in the process that inadvertently inject bias. Hence, we hold the view that bias is not an inherent property of the data itself; how we treat the data during different phases of the annotation project is what introduces bias. In line with this view, we consider those intrinsic attributes as only the sources of bias---not the bias itself. In this subsection, we briefly review what might constitute the sources of bias in the data in an annotation project. Then, in the following subsections we address how the team should remedy the biases and avoid the pitfalls.

        First, OoIs, as the natural subjects of the study (1) may have a skewed distribution (imbalanced), making it difficult to sample a large-enough number of their instances. (2) They may also be too small, too large, or of varying sizes, rendering some of the annotation expectations impractical. (3) The cycles in which their potentially reoccurring patterns should be discovered might be too long (longer than the data collection period itself) or unknown. Similarly, (4) the regions the collected data spans may be a subset of the regions from which representative data should have been sampled from. Depending on the specific subjects, other similar sources of bias can be identified.

        Second, the sensory devices which produce the image data may be the sources of bias as well. For instance, (1) the utilized cameras may have inconsistent spatial resolutions, resulting in inconsistent analytics. (2) The sensory devices may produce different image formats, needing a unifying conversion algorithm, which consequently may leave some formats to lose more information than others. (3) The devices, subject to degradation, may be of different ages, and therefore, produce images of different qualities. (4) The superior location of some devices (due to better lighting, humidity, temperature, etc.) may also result in inconsistent qualities across devices. (5) Upholding different maintenance protocols can also vary the image qualities from one device to another. These are just a few examples of how sensory devices can introduce bias.

        Third, the strategies and routines by which the sensory devices collect data could also form sources of bias. To name a few, the cadence at which each sensory device collects samples, individually, or together with other devices, may be insufficient to capture some natural patterns. Moreover, the OoIs instances which devices encounter, owing to the devices' placement and timing of data collection, might not be uniform. A significant illustration of this scenario is medical imaging conducted in a health center where the facility's location determines a demographic (or possibly a specific socioeconomic group) that does not reflect the full spectrum of human medical diversity. Depending on how the collected data might be used, this may be understood as bias or specifics of the data.
        
        Since it is merely impossible for us to provide a comprehensive list of ways by which bias may permeate annotation we use an example instead, to illustrate the complexities associated with identifying a source of bias, hoping that it helps the team identify similar patterns in their own project. Vigor maps are created by flying unmanned aerial vehicles (UAVs) over crop fields \citep{lelong2008assessment}. Such maps are used in precision agriculture to examine the vitality and health of crops \citep{sishodia2020applications}. Presence of clouds which are more frequent in some seasons, is an important obstacle in aerial imaging. To mitigate their effect on the quality of images, aerial passes can be scheduled according to the presence of clouds. However, whether such appropriate strategies have been employed or not is beyond the control of the annotation team (see the difference between data collection and data sampling in Fig.~\ref{fig:data_source_pool_feed_pipeline}). Nevertheless, a lack of awareness about these sources of bias leads to their infiltration into the annotation pipeline. For example, the annotation team might exclude a large collection of aerial images taken during certain seasons to avoid low-quality ('cloudy') images. Consequently, the manually annotated dataset will inherit the bias of non-representative samples, culminating in a biased annotated dataset.

    \subsection{Bias in Sampling}\label{subsec:bias-in-sampling}
        
        One of the objectives of the sampling strategies, as discussed in Sec.~\ref{subsec:data-sampling}, is to compensate for any bias that is already present in the data source (stemming from the adopted data collection strategies), and to minimize the introduction of any new bias. Below, we list a few of the most general ones.
    
        When sampling does not follow the expected distribution of the image sources: The image sources may vary in terms of the types of imaging instruments used, the configurations of the instruments, or the observation setting (lighting, exposure, temperature, humidity, etc.) in which they are used. Also, the expected distribution of image sources may be the same as, or different from, the actual distribution of image sources in the data. Having similar distributions preserves the reality, while for the sake of an effective training of an machine learning model, one image source may need to be sampled more frequently. For example, suppose a subset of images are sampled from multiple devices (e.g., MRI machines with open or closed configurations; telescopes with different apertures or at different locations) and the end goal of the annotation project is to train a supervised model that automatically annotates the images captured by any of those devices. If only a small portion of the images are sampled from one device, the trained AI model is likely to under-perform on images sampled from that source due to lack of proportional representation.
        
        
        When sampling does not follow the expected distribution of OoIs: Nature does not guarantee a uniform distribution of OoIs; one OoI may occur more frequently than others (e.g., satellite images depicting the presence or absence of oil spills in the oceans). However, since machine learning algorithms perform least biased when trained on balanced datasets, often a uniform distribution of OoIs needs to be sent over for annotation. This is regardless of how this was treated previously in the data collection phase (beyond the control of the annotation team). An unbiased sampling strategy may be designed to sample OoIs uniformly despite their (natural) non-uniform distributions. There does not always exist a simple approach for such a requirement. Such a sampling strategy may depend on some knowledge about where the rare events can be found; locating the rare events (e.g., oil spills) could be after all the very task for which the manual annotation is to be carried out. Lack of a solution for this problem may result in biased sampled data.
    
        When sampling does not reflect the long-term trends of OoIs in the data: This is a particular point of caution for OoIs with important spatial and/or temporal characteristics. A biased sampling strategy may collect fewer images from certain periods of time, or certain locations. For example, vigor maps are created from aerial imaging and are used in precision agriculture to examine the vitality and health of crops \citep{sishodia2020applications}. Presence of clouds which are more frequent in some seasons, is an important obstacle in aerial imaging. To mitigate their effect, aerial passes are scheduled according to the presence of clouds. However, an image annotation task usually relies on an existing data source and cannot manipulate field data collection strategies. Therefore, a sampling strategy that is indifferent to such bias lets the embedded bias pass through the annotation pipeline.

    \subsection{Bias in Data Distribution}\label{subsec:bias-in-data-distribution}
        Data distribution strategies can also introduce bias into the project. This can occur either by implementing new biased strategies or by failing to account for the biases introduced previously by the collection and/or sampling strategies. A non-representative distribution strategy inherently carries bias. As the name suggests, such a strategy does not preserve some natural properties of the OoIs. A few of such properties (although, not an exclusive list) are (1) prevalence of one OoI compared to that of others (e.g., instances of one species, in the distributed images, appear more frequently in the distributed images whether this imbalance is a natural phenomenon or not, or (2) the location of one OoI compared to that of the others (e.g., when breast cancer cells, in the distributed images, are more frequently observed near the lymph nodes without the images being associated to ``lymph-node positive'' cases which is only expected in 26\% of the relevant patients \citep{lou2023incidence}), or (3) the color, shape, or texture of an OoI compared to that of others (e.g., solar filaments with dextral chirality, in the distributed images, appear wider or thinner compared to those with sinistral chirality, which is not a known natural property of filaments in Heliophysics). Not taking into account such properties in the design of the data distribution strategies could make the annotators follow superficial patterns. Note that due to the repetitive nature of the manual annotation task, the annotators constantly develop simplifying rules---consciously or subconsciously---for identifying the OoI instances and for determining their PoIs. Since, as discussed earlier, the annotation of scientific OoIs requires more work and critical thinking, the desire for simplification is even stronger. Such superficial patterns are very likely to be assumed by the annotators as factual, and therefore, form biased annotated data.

        Additionally, the distribution strategies may lead to an uneven distribution of workload among the annotators, which can consequently complicate the fair compensation of their work. We already discussed this in Sec.~\ref{subsec:data-distribution-strategy}. Note that the distribution strategies utilized in the data feed pipeline are the main place where this issue can be addressed unless the payment method employed to compensate the annotators' work relies on measures other than the number of the annotated images or OoI instances.

        Further, recall the aggregation strategies we listed in Sec.~\ref{subsec:data-distribution-strategy}. In the best-overrides-rest method, the notion of `best' is subjective, therefore, can potentially introduce bias. As an example, suppose that the intersection operation is used for aggregation of segmentations associated to each OoI instance. This decision will result in having systematically smaller shapes than the smallest shape that was manually created. Conversely, using the union operation will result in bigger PoI instances. More importantly, for annotation tasks in which a high granularity of segmentations is expected from the annotators, either of these operations (union or intersection) will significantly reduce the granularities which the annotators spent a lot of time to carefully create. The annotation team should identify such sources of bias when designing the data distribution strategies.

    \subsection{Bias in Training and Training Material}
        Generally, to minimize bias in training, the team should (1) focus on an objective training and (2) encourage neutrality. An objective training constitutes providing the annotators with clear, objective criteria for identifying and annotating OoIs, without mentioning untested theories (particularly those which the annotations are intended to be used for) or expected outcomes (if already known). To encourage neutrality, the trainers should stress the importance of unbiased observations and annotations, ensuring annotators document what they see rather than what they expect to see. It is a good idea, if time permits, to educate the annotators on how the relevant forms of bias can manifest themselves in this particular project. Different types of bias are discussed in Sec.~\ref{subsec:bias-in-annotation}.

        The training material should be reviewed to assure the neutrality tone in the language and graphics. There are numerous places where a training resource may inadvertently suggest a biased practice. For example, in an annotation guide the OoIs may be listed in a way that places more importance over some labels (e.g., by simply providing a more detailed definition for some OoIs). The annotators who are led to believe in such a superficial ranking of OoIs may be less careful when annotating the ``less important'' OoI instances, resulting in an uneven precision in the annotations.

        The trainers or the training material may give away pieces of information which can potentially impact the annotators' objectiveness. For example, it might be known to the domain experts that a particular type of tumor constitutes about 10\% of the OoIs in a given dataset. Exposing the annotators to this statistic may let them alter their decisions when labeling tissues, solely to comply with the ``expected'' one-to-nine ratio. This is generally known as the confirmation bias \citep{nickerson1998confirmation} and should be minimized to the extent possible.

        Any unrealistic expectations set by the trainers or the training materials can compel annotators to resort to ad hoc solutions to meet the requirements. For example, if during the training sessions the annotators are shown how to capture some complex structures of the OoIs, but the expected time for annotating each instance is much less than the time needed to capture such complex structures, they may naturally attempt to incorporate some superficial granularities in their annotations only to mimic what a proper annotation should look like. Such superficial granularities may be very difficult to catch (and then correct) during the review process. This would result in either a vicious (and expensive) cycle of returning the annotations to the annotators for improvements, or a dataset of annotations with superficial details. Such a bias (generally known as the Dunning-Kruger effect \citep{kruger1999unskilled}) is extremely likely to take place if the team does not thoroughly and empirically test the difficulty of meeting the defined expectations, in a realistic setting.

        Good trainers equip the annotators with some tricks and rule-of-thumbs to enhance the quality and quantity in the annotation process. While those tricks are generally very helpful and innocent (such as ignoring OoIs which are smaller than $n$ pixels long), some others may form biased opinions among the annotators. For instance, the annotators may be trained to consider the possibility that spatially-close OoI instances might be in fact a single instance appearing as multiple instances. Without providing further criteria for testing such a possibility the annotators may, despite any strong evidence, identify multiple spatially-close objects as a single OoI instance. They may even go further and alter their initial assignment of OoI only to justify the formation of the bigger OoI instance. It is best that those tricks and rule-of-thumbs are consulted with the rest of the team to ensure their potential impact.

    \subsection{Bias in Annotation}\label{subsec:bias-in-annotation}
        Annotators bring human bias to the equation. Human bias in repetitive tasks such as manual annotation manifest itself in different forms of which we list a few \citep{tejani2024understanding}. (1) The \textit{confirmation bias} is the tendency to follow paths which agree with our prior beliefs. When annotators are faced with OoI instances which are hard to discern (often because they exhibit properties of two or more classes), they tend to favor patterns which agree with their initial judgment. (2) The availability bias manifest itself by increasing the chance of assigning a label depending on what labels was used recently. This becomes more prominent (overlapping with the expectation bias \citep{rosenthal1968pygmalion}) when an annotator develops a strategy to annotate OoIs one class at a time. (3) The anchoring bias refers to the tendency to refuse to change the initial judgement despite contrary evidence. This occurs when the annotator fails to notice clear patterns during the creation of PoIs that contradict the initially assigned labels.

        The reviewers also carry their own biases. If the review process allows the reviewers to go over each annotator's work separately, they quickly develop judgment on the quality of their annotations. This judgement is valuable and can be used to provide constructive and holistic feedback for the annotator, helping them improve their work. However, the reviewer's confirmation bias may cause inaccurate annotations go unnoticed, or conversely, acceptable annotations be judged too harshly. A reviewer may also exhibit anchoring bias. It is important to remember that the review process can be almost as tedious as the annotation itself, and humans tend to avoid this effort whenever possible. Therefore, there is a tendency to agree with the reviewers' annotations.

        Being aware of how human bias can permeate the annotation process, the team should develop practices to mitigate them to the extent possible. They can provide an exclusive set of labels (including the `in-between' cases) to allow the annotators choose more confidently. Note that such labels can be used for the sole purpose of catching the difficult cases and working on them separately. The team can also provide explicit examples of difficult cases and illustrate the criteria based on which the decision should be made. This is where the supplemental sources of information can be helpful (see Sec.~\ref{subsec:supplemental}). Automated cross-verification methods can also be developed, not for the aggregation of the labels, but to identify annotations with lower confidence (e.g., with high discrepancies among the annotators of the same group) and tag them for further review.

\section{Roles to Define}\label{sec:roles}
    The size of the annotation team depends on the size of the annotation project, which is somewhat determined by the number of images targeted to be annotated. But the roles in the team remain---more or less---the same, regardless. The only difference is that in smaller teams each member is expected to wear multiple hats. In this section, we review the main roles.
    
    \textit{Team leaders} are the main organizers of the project and sit at the top of the pyramid of responsibilities. They oversee the entire project, both the technical and the administrative aspects of the project.
    
    \textit{Domain experts} are responsible for (1) giving the team necessary consultation on the subject of study, the value of the included OoIs and PoIs, the details about the instruments which produce the image data, and all the little things without which the annotation project will lose some quality, (2) taking part or leading the annotation training sessions, (3) taking part or leading the annotation review process, and (4) identifying potential mistakes by being engaged in the documentation of the project. It is crucial that all team members understand that domain experts are not merely occasional guests in the meetings; they must be engaged in every single decision made by the team from the very beginning when the team learn about the data, to the very end when the post-processed annotations are to be disseminated for use by the targeted community. This required level of engagement may also be used to determine who can serve this role the best. 
    
    \textit{Data scientists} are responsible for (1) identifying the unique (and potentially challenging) features of the data, (2) providing the team with data analytics which help design the data sampling and integration strategies, (3) providing strategies for the preparation of data (e.g., format conversion, homogenization, etc.), (4) identifying the biases and designing validation measures for monitoring the quality of the manual annotation process during and after the annotation phase. Data scientists are the communication channel between the domain experts and the rest of the team. Therefore, the most important criterion for choosing the right team of data scientists is arguably their interdisciplinary experience and understanding of the intricacies involved in effective communication among team members within an interdisciplinary setting.
    
    \textit{Data engineers} are responsible for (1) choosing the software stack and services needed during the entire project period, (2) implementing the code for data sampling, preparation, and integration, (3) developing the data distribution strategy and the data feed pipeline, (4) providing the training data for the annotation training sessions, and (5) performing post-processing, storage, and dissemination of the annotated data.
    
    \textit{Reviewers} and \textit{Expert Reviewers} are responsible for: (1) reviewing the annotators' work (accepting or rejecting annotation, and providing the annotators with constructive feedback), (2) directly improving the annotations' quality when needed, and (3) maintaining a communication channel between the annotators and the rest of the team to ensure that their challenges are addressed.
    
    \textit{Annotators} are responsible for the actual annotation around which the entire project revolves. By breaking down this role into sub-roles (e.g., \textit{annotators}, \textit{annotation admins}, \textit{modifiers}, etc.), a more advanced and more effective annotation workflow can be built.
    To maximize the quality of their work, different strategies may be adopted, but the most important of all is to maintain an open communication channel and make sure that they feel comfortable casting their opinion on the existing challenges. This is primarily because the annotators are the only group which directly create the annotations, and therefore, the issues which they might notice may be very difficult (or too late) for the other team members to identify.
    
    Last, but certainly not least, \textit{resource managers} are responsible for (1) purchasing the services, tools, and technologies, (2) hiring the new annotators, (3) processing the payments in a timely manner, and (4) keeping the project on the budget. In the absence of a dedicated resource manager, this burden may fall on the team leader and this can potentially hinder the project and/or impact the quality of the final product.

\section{Conclusion}
    We put together a guide paper underscoring the complexity and necessity of well-managed manual annotation projects for scientific images. By addressing the multifaceted challenges such as data collection, resource allocation, recruitment, and bias mitigation, we offer a domain-agnostic guide to streamline these efforts. Our insights, drawn from extensive experience, highlight key aspects like defining success measures, understanding annotation subjects, clarifying project goals, assessing data availability, and identifying the needs in the team. Additionally, we emphasize the importance of leveraging appropriate tools and technologies to enhance both the quality and efficiency of annotations. Our aim is to foster further research and collaboration, ultimately contributing to a robust body of knowledge that will lower the overall costs and improve the outcomes of manual-annotation projects across diverse scientific fields.

\section*{Acknowledgement}
    This material is based upon work supported by the National Science Foundation under Grant No. 2209912 and 2433781, directorate for Computer and Information Science and Engineering (CSE), and Office of Advanced Cyberinfrastructure (OAC).




\bibliographystyle{elsarticle-harv}
\bibliography{main.bib}

\begin{thebibliography}{68}
\expandafter\ifx\csname natexlab\endcsname\relax\def\natexlab#1{#1}\fi
\providecommand{\url}[1]{\texttt{#1}}
\providecommand{\href}[2]{#2}
\providecommand{\path}[1]{#1}
\providecommand{\DOIprefix}{doi:}
\providecommand{\ArXivprefix}{arXiv:}
\providecommand{\URLprefix}{URL: }
\providecommand{\Pubmedprefix}{pmid:}
\providecommand{\doi}[1]{\href{http://dx.doi.org/#1}{\path{#1}}}
\providecommand{\Pubmed}[1]{\href{pmid:#1}{\path{#1}}}
\providecommand{\bibinfo}[2]{#2}
\ifx\xfnm\relax \def\xfnm[#1]{\unskip,\space#1}\fi
\bibitem[{Acharya and Ray(2005)}]{acharya2005image}
\bibinfo{author}{Acharya, T.}, \bibinfo{author}{Ray, A.K.}, \bibinfo{year}{2005}.
\newblock \bibinfo{title}{Image processing: principles and applications}.
\newblock \bibinfo{publisher}{John Wiley \& Sons}.
\bibitem[{Ahmadzadeh et~al.(2019)Ahmadzadeh, Kempton and Angryk}]{Ahmadzadeh2019curated}
\bibinfo{author}{Ahmadzadeh, A.}, \bibinfo{author}{Kempton, D.J.}, \bibinfo{author}{Angryk, R.A.}, \bibinfo{year}{2019}.
\newblock \bibinfo{title}{A curated image parameter data set from the solar dynamics observatory mission}.
\newblock \bibinfo{journal}{The Astrophysical Journal Supplement Series} \bibinfo{volume}{243}, \bibinfo{pages}{18}.
\newblock \URLprefix \url{https://dx.doi.org/10.3847/1538-4365/ab253a}, \DOIprefix\doi{10.3847/1538-4365/ab253a}.
\bibitem[{Ahmadzadeh et~al.(2021)Ahmadzadeh, Kempton, Chen and Angryk}]{ahmadzadeh2021multiscale}
\bibinfo{author}{Ahmadzadeh, A.}, \bibinfo{author}{Kempton, D.J.}, \bibinfo{author}{Chen, Y.}, \bibinfo{author}{Angryk, R.A.}, \bibinfo{year}{2021}.
\newblock \bibinfo{title}{Multiscale iou: {A} metric for evaluation of salient object detection with fine structures}.
\newblock \bibinfo{journal}{CoRR} \bibinfo{volume}{abs/2105.14572}.
\newblock \URLprefix \url{https://arxiv.org/abs/2105.14572}, \href{http://arxiv.org/abs/2105.14572}{{\tt arXiv:2105.14572}}.
\bibitem[{Ampatzoglou et~al.(2016)Ampatzoglou, Ampatzoglou, Chatzigeorgiou, Avgeriou, Abrahamsson, Martini, Zdun and Systa}]{ampatzoglou2016perception}
\bibinfo{author}{Ampatzoglou, A.}, \bibinfo{author}{Ampatzoglou, A.}, \bibinfo{author}{Chatzigeorgiou, A.}, \bibinfo{author}{Avgeriou, P.}, \bibinfo{author}{Abrahamsson, P.}, \bibinfo{author}{Martini, A.}, \bibinfo{author}{Zdun, U.}, \bibinfo{author}{Systa, K.}, \bibinfo{year}{2016}.
\newblock \bibinfo{title}{The perception of technical debt in the embedded systems domain: An industrial case study}, in: \bibinfo{booktitle}{2016 IEEE 8th International Workshop on Managing Technical Debt (MTD)}, pp. \bibinfo{pages}{9--16}.
\newblock \DOIprefix\doi{10.1109/MTD.2016.8}.
\bibitem[{Barnsley et~al.(1988)Barnsley, Devaney, Mandelbrot, Peitgen, Saupe, Voss, Fisher and McGuire}]{barnsley1988science}
\bibinfo{author}{Barnsley, M.F.}, \bibinfo{author}{Devaney, R.L.}, \bibinfo{author}{Mandelbrot, B.B.}, \bibinfo{author}{Peitgen, H.O.}, \bibinfo{author}{Saupe, D.}, \bibinfo{author}{Voss, R.F.}, \bibinfo{author}{Fisher, Y.}, \bibinfo{author}{McGuire, M.}, \bibinfo{year}{1988}.
\newblock \bibinfo{title}{The science of fractal images}.
\newblock \bibinfo{publisher}{Springer}.
\bibitem[{Boykov and Jolly(2001)}]{boykov2001interactive}
\bibinfo{author}{Boykov, Y.}, \bibinfo{author}{Jolly, M.P.}, \bibinfo{year}{2001}.
\newblock \bibinfo{title}{Interactive graph cuts for optimal boundary \& region segmentation of objects in n-d images}, in: \bibinfo{booktitle}{Proceedings Eighth IEEE International Conference on Computer Vision. ICCV 2001}, pp. \bibinfo{pages}{105--112 vol.1}.
\newblock \DOIprefix\doi{10.1109/ICCV.2001.937505}.
\bibitem[{Branson et~al.(2017)Branson, Van~Horn and Perona}]{Branson2017lean}
\bibinfo{author}{Branson, S.}, \bibinfo{author}{Van~Horn, G.}, \bibinfo{author}{Perona, P.}, \bibinfo{year}{2017}.
\newblock \bibinfo{title}{Lean crowdsourcing: Combining humans and machines in an online system}, in: \bibinfo{booktitle}{2017 IEEE Conference on Computer Vision and Pattern Recognition (CVPR)}, pp. \bibinfo{pages}{6109--6118}.
\newblock \DOIprefix\doi{10.1109/CVPR.2017.647}.
\bibitem[{{Bryan C. Russell, Antonio Torralba, Kevin P. Murphy, William T. Freeman}(2024)}]{labelme}
\bibinfo{author}{{Bryan C. Russell, Antonio Torralba, Kevin P. Murphy, William T. Freeman}}, \bibinfo{year}{2024}.
\newblock \bibinfo{title}{Labelme}.
\newblock \bibinfo{howpublished}{\url{https://github.com/labelmeai/labelme}}.
\newblock \bibinfo{note}{[Online; accessed 01-June-2024]}.
\bibitem[{Bustos et~al.(2020)Bustos, Pertusa, Salinas and {de la Iglesia-Vayá}}]{bustos2020padchest}
\bibinfo{author}{Bustos, A.}, \bibinfo{author}{Pertusa, A.}, \bibinfo{author}{Salinas, J.M.}, \bibinfo{author}{{de la Iglesia-Vayá}, M.}, \bibinfo{year}{2020}.
\newblock \bibinfo{title}{Padchest: A large chest x-ray image dataset with multi-label annotated reports}.
\newblock \bibinfo{journal}{Medical Image Analysis} \bibinfo{volume}{66}, \bibinfo{pages}{101797}.
\newblock \URLprefix \url{https://www.sciencedirect.com/science/article/pii/S1361841520301614}, \DOIprefix\doi{https://doi.org/10.1016/j.media.2020.101797}.
\bibitem[{Dave et~al.(2015)Dave, Joshi and Srivastava}]{dave2015survey}
\bibinfo{author}{Dave, C.P.}, \bibinfo{author}{Joshi, R.}, \bibinfo{author}{Srivastava, S.}, \bibinfo{year}{2015}.
\newblock \bibinfo{title}{A survey on geometric correction of satellite imagery}.
\newblock \bibinfo{journal}{International Journal of Computer Applications} \bibinfo{volume}{116}.
\bibitem[{Deng et~al.(2009)Deng, Dong, Socher, Li, Li and Fei-Fei}]{deng2009imagenet}
\bibinfo{author}{Deng, J.}, \bibinfo{author}{Dong, W.}, \bibinfo{author}{Socher, R.}, \bibinfo{author}{Li, L.J.}, \bibinfo{author}{Li, K.}, \bibinfo{author}{Fei-Fei, L.}, \bibinfo{year}{2009}.
\newblock \bibinfo{title}{Imagenet: A large-scale hierarchical image database}, in: \bibinfo{booktitle}{2009 IEEE Conference on Computer Vision and Pattern Recognition}, pp. \bibinfo{pages}{248--255}.
\newblock \DOIprefix\doi{10.1109/CVPR.2009.5206848}.
\bibitem[{Donlon et~al.(2021)Donlon, Cullen, Giulicchi, Vuilleumier, Francis, Kuschnerus, Simpson, Bouridah, Caleno, Bertoni, Rancaño, Pourier, Hyslop, Mulcahy, Knockaert, Hunter, Webb, Fornari, Vaze, Brown, Willis, Desai, Desjonqueres, Scharroo, Martin-Puig, Leuliette, Egido, Smith, Bonnefond, {Le Gac}, Picot and Tavernier}]{donlon2021copernicus}
\bibinfo{author}{Donlon, C.J.}, \bibinfo{author}{Cullen, R.}, \bibinfo{author}{Giulicchi, L.}, \bibinfo{author}{Vuilleumier, P.}, \bibinfo{author}{Francis, C.R.}, \bibinfo{author}{Kuschnerus, M.}, \bibinfo{author}{Simpson, W.}, \bibinfo{author}{Bouridah, A.}, \bibinfo{author}{Caleno, M.}, \bibinfo{author}{Bertoni, R.}, \bibinfo{author}{Rancaño, J.}, \bibinfo{author}{Pourier, E.}, \bibinfo{author}{Hyslop, A.}, \bibinfo{author}{Mulcahy, J.}, \bibinfo{author}{Knockaert, R.}, \bibinfo{author}{Hunter, C.}, \bibinfo{author}{Webb, A.}, \bibinfo{author}{Fornari, M.}, \bibinfo{author}{Vaze, P.}, \bibinfo{author}{Brown, S.}, \bibinfo{author}{Willis, J.}, \bibinfo{author}{Desai, S.}, \bibinfo{author}{Desjonqueres, J.D.}, \bibinfo{author}{Scharroo, R.}, \bibinfo{author}{Martin-Puig, C.}, \bibinfo{author}{Leuliette, E.}, \bibinfo{author}{Egido, A.}, \bibinfo{author}{Smith, W.H.}, \bibinfo{author}{Bonnefond, P.}, \bibinfo{author}{{Le Gac}, S.}, \bibinfo{author}{Picot, N.}, \bibinfo{author}{Tavernier, G.},
  \bibinfo{year}{2021}.
\newblock \bibinfo{title}{The copernicus sentinel-6 mission: Enhanced continuity of satellite sea level measurements from space}.
\newblock \bibinfo{journal}{Remote Sensing of Environment} \bibinfo{volume}{258}, \bibinfo{pages}{112395}.
\newblock \URLprefix \url{https://www.sciencedirect.com/science/article/pii/S0034425721001139}, \DOIprefix\doi{https://doi.org/10.1016/j.rse.2021.112395}.
\bibitem[{Dutta et~al.(2016)Dutta, Gupta and Zissermann}]{vggImageAnnotator}
\bibinfo{author}{Dutta, A.}, \bibinfo{author}{Gupta, A.}, \bibinfo{author}{Zissermann, A.}, \bibinfo{year}{2016}.
\newblock \bibinfo{title}{{VGG} image annotator ({VIA})}.
\newblock \bibinfo{howpublished}{\url{http://www.robots.ox.ac.uk/~vgg/software/via/}}.
\newblock \bibinfo{note}{Online; accessed 01-June-2024}.
\bibitem[{Eastman et~al.(2011)Eastman, Netanyahu and Moigne}]{eastmansurvey2011}
\bibinfo{author}{Eastman, R.D.}, \bibinfo{author}{Netanyahu, N.S.}, \bibinfo{author}{Moigne, J.L.}, \bibinfo{year}{2011}.
\newblock \bibinfo{title}{Survey of image registration methods}, in: \bibinfo{editor}{Moigne, J.L.}, \bibinfo{editor}{Netanyahu, N.S.}, \bibinfo{editor}{Eastman, R.D.} (Eds.), \bibinfo{booktitle}{Image Registration for Remote Sensing}. \bibinfo{publisher}{Cambridge University Press}, pp. \bibinfo{pages}{35--78}.
\bibitem[{{Emily Mason}(2022)}]{citizenscience2022ARSpotter}
\bibinfo{author}{{Emily Mason}, K.}, \bibinfo{year}{2022}.
\newblock \bibinfo{title}{Citizen science - solar active region spotter}.
\newblock \bibinfo{howpublished}{\url{https://www.zooniverse.org/projects/eimason/solar-active-region-spotter}}.
\newblock \bibinfo{note}{[Online; accessed 13-June-2023]}.
\bibitem[{Everingham et~al.(2010)Everingham, Van~Gool, Williams, Winn and Zisserman}]{everingham2010pascal}
\bibinfo{author}{Everingham, M.}, \bibinfo{author}{Van~Gool, L.}, \bibinfo{author}{Williams, C.K.}, \bibinfo{author}{Winn, J.}, \bibinfo{author}{Zisserman, A.}, \bibinfo{year}{2010}.
\newblock \bibinfo{title}{The pascal visual object classes (voc) challenge}.
\newblock \bibinfo{journal}{International journal of computer vision} \bibinfo{volume}{88}, \bibinfo{pages}{303--338}.
\bibitem[{Ghanavati et~al.(2014)Ghanavati, Lerch and Sled}]{ghanavati2014automatic}
\bibinfo{author}{Ghanavati, S.}, \bibinfo{author}{Lerch, J.P.}, \bibinfo{author}{Sled, J.G.}, \bibinfo{year}{2014}.
\newblock \bibinfo{title}{Automatic anatomical labeling of the complete cerebral vasculature in mouse models}.
\newblock \bibinfo{journal}{NeuroImage} \bibinfo{volume}{95}, \bibinfo{pages}{117--128}.
\newblock \URLprefix \url{https://www.sciencedirect.com/science/article/pii/S1053811914002043}, \DOIprefix\doi{https://doi.org/10.1016/j.neuroimage.2014.03.044}.
\bibitem[{Gupta et~al.(2019)Gupta, Dollar and Girshick}]{Gupta2019lvis}
\bibinfo{author}{Gupta, A.}, \bibinfo{author}{Dollar, P.}, \bibinfo{author}{Girshick, R.}, \bibinfo{year}{2019}.
\newblock \bibinfo{title}{Lvis: A dataset for large vocabulary instance segmentation}, in: \bibinfo{booktitle}{Proceedings of the IEEE/CVF Conference on Computer Vision and Pattern Recognition (CVPR)}, pp. \bibinfo{pages}{5356--5364}.
\bibitem[{Hajnal and Hill(2001)}]{hajnal2001medical}
\bibinfo{author}{Hajnal, J.V.}, \bibinfo{author}{Hill, D.L.}, \bibinfo{year}{2001}.
\newblock \bibinfo{title}{Medical image registration}.
\newblock \bibinfo{publisher}{CRC press}.
\newblock \DOIprefix\doi{https://doi.org/10.1201/9781420042474}.
\bibitem[{Hao(2023)}]{Hao_2023}
\bibinfo{author}{Hao, K.}, \bibinfo{year}{2023}.
\newblock \bibinfo{title}{How the ai industry profits from catastrophe}.
\newblock \URLprefix \url{https://www.technologyreview.com/2022/04/20/1050392/ai-industry-appen-scale-data-labels/}.
\bibitem[{Kirillov et~al.(2023)Kirillov, Mintun, Ravi, Mao, Rolland, Gustafson, Xiao, Whitehead, Berg, Lo et~al.}]{kirillov2023segment}
\bibinfo{author}{Kirillov, A.}, \bibinfo{author}{Mintun, E.}, \bibinfo{author}{Ravi, N.}, \bibinfo{author}{Mao, H.}, \bibinfo{author}{Rolland, C.}, \bibinfo{author}{Gustafson, L.}, \bibinfo{author}{Xiao, T.}, \bibinfo{author}{Whitehead, S.}, \bibinfo{author}{Berg, A.C.}, \bibinfo{author}{Lo, W.Y.}, et~al., \bibinfo{year}{2023}.
\newblock \bibinfo{title}{Segment anything}.
\newblock \bibinfo{journal}{arXiv preprint arXiv:2304.02643} .
\bibitem[{Krizhevsky et~al.(2009)Krizhevsky, Hinton et~al.}]{krizhevsky2009learning}
\bibinfo{author}{Krizhevsky, A.}, \bibinfo{author}{Hinton, G.}, et~al., \bibinfo{year}{2009}.
\newblock \bibinfo{title}{Learning multiple layers of features from tiny images}.
\newblock \bibinfo{journal}{na} \bibinfo{volume}{na}.
\newblock \URLprefix \url{https://www.cs.utoronto.ca/~kriz/learning-features-2009-TR.pdf}.
\bibitem[{Kruchten et~al.(2012)Kruchten, Nord and Ozkaya}]{kruchten2012technical}
\bibinfo{author}{Kruchten, P.}, \bibinfo{author}{Nord, R.L.}, \bibinfo{author}{Ozkaya, I.}, \bibinfo{year}{2012}.
\newblock \bibinfo{title}{Technical debt: From metaphor to theory and practice}.
\newblock \bibinfo{journal}{IEEE Software} \bibinfo{volume}{29}, \bibinfo{pages}{18--21}.
\newblock \DOIprefix\doi{10.1109/MS.2012.167}.
\bibitem[{Kruger and Dunning(1999)}]{kruger1999unskilled}
\bibinfo{author}{Kruger, J.}, \bibinfo{author}{Dunning, D.}, \bibinfo{year}{1999}.
\newblock \bibinfo{title}{Unskilled and unaware of it: how difficulties in recognizing one's own incompetence lead to inflated self-assessments.}
\newblock \bibinfo{journal}{Journal of personality and social psychology} \bibinfo{volume}{77}, \bibinfo{pages}{1121}.
\newblock \URLprefix \url{https://doi.org/10.1037/0022-3514.77.6.1121}, \DOIprefix\doi{10.1037/0022-3514.77.6.1121}.
\bibitem[{Kuznetsova et~al.(2020)Kuznetsova, Rom, Alldrin, Uijlings, Krasin, Pont-Tuset, Kamali, Popov, Malloci, Kolesnikov et~al.}]{kuznetsova2020open}
\bibinfo{author}{Kuznetsova, A.}, \bibinfo{author}{Rom, H.}, \bibinfo{author}{Alldrin, N.}, \bibinfo{author}{Uijlings, J.}, \bibinfo{author}{Krasin, I.}, \bibinfo{author}{Pont-Tuset, J.}, \bibinfo{author}{Kamali, S.}, \bibinfo{author}{Popov, S.}, \bibinfo{author}{Malloci, M.}, \bibinfo{author}{Kolesnikov, A.}, et~al., \bibinfo{year}{2020}.
\newblock \bibinfo{title}{The open images dataset v4: Unified image classification, object detection, and visual relationship detection at scale}.
\newblock \bibinfo{journal}{International Journal of Computer Vision} \bibinfo{volume}{128}, \bibinfo{pages}{1956--1981}.
\newblock \DOIprefix\doi{https://doi.org/10.1007/s11263-020-01316-z}.
\bibitem[{LeCun(1998)}]{lecun1998mnist}
\bibinfo{author}{LeCun, Y.}, \bibinfo{year}{1998}.
\newblock \bibinfo{title}{The mnist database of handwritten digits}.
\newblock \bibinfo{journal}{http://yann. lecun. com/exdb/mnist/} .
\bibitem[{LeCun(2019)}]{lecun2019deep}
\bibinfo{author}{LeCun, Y.}, \bibinfo{year}{2019}.
\newblock \bibinfo{title}{1.1 deep learning hardware: Past, present, and future}, in: \bibinfo{booktitle}{2019 IEEE International Solid-State Circuits Conference - (ISSCC)}, pp. \bibinfo{pages}{12--19}.
\newblock \DOIprefix\doi{10.1109/ISSCC.2019.8662396}.
\bibitem[{Lelong et~al.(2008)Lelong, Burger, Jubelin, Roux, Labbé and Baret}]{lelong2008assessment}
\bibinfo{author}{Lelong, C.C.D.}, \bibinfo{author}{Burger, P.}, \bibinfo{author}{Jubelin, G.}, \bibinfo{author}{Roux, B.}, \bibinfo{author}{Labbé, S.}, \bibinfo{author}{Baret, F.}, \bibinfo{year}{2008}.
\newblock \bibinfo{title}{Assessment of unmanned aerial vehicles imagery for quantitative monitoring of wheat crop in small plots}.
\newblock \bibinfo{journal}{Sensors} \bibinfo{volume}{8}, \bibinfo{pages}{3557--3585}.
\newblock \URLprefix \url{https://www.mdpi.com/1424-8220/8/5/3557}, \DOIprefix\doi{10.3390/s8053557}.
\bibitem[{Liao et~al.(2021)Liao, Kar and Fidler}]{Liao2021towards}
\bibinfo{author}{Liao, Y.H.}, \bibinfo{author}{Kar, A.}, \bibinfo{author}{Fidler, S.}, \bibinfo{year}{2021}.
\newblock \bibinfo{title}{Towards good practices for efficiently annotating large-scale image classification datasets}, in: \bibinfo{booktitle}{Proceedings of the IEEE/CVF Conference on Computer Vision and Pattern Recognition (CVPR)}, pp. \bibinfo{pages}{4350--4359}.
\bibitem[{Lin(2014)}]{code2014cocoapi}
\bibinfo{author}{Lin, T.Y.}, \bibinfo{year}{2014}.
\newblock \bibinfo{title}{Ms coco api}.
\newblock \bibinfo{howpublished}{\url{https://github.com/cocodataset/cocoapi}}.
\newblock \bibinfo{note}{[Online; accessed 13-June-2023]}.
\bibitem[{Lin et~al.(2014)Lin, Maire, Belongie, Hays, Perona, Ramanan, Doll{\'a}r and Zitnick}]{lin2014coco}
\bibinfo{author}{Lin, T.Y.}, \bibinfo{author}{Maire, M.}, \bibinfo{author}{Belongie, S.}, \bibinfo{author}{Hays, J.}, \bibinfo{author}{Perona, P.}, \bibinfo{author}{Ramanan, D.}, \bibinfo{author}{Doll{\'a}r, P.}, \bibinfo{author}{Zitnick, C.L.}, \bibinfo{year}{2014}.
\newblock \bibinfo{title}{Microsoft coco: Common objects in context}, in: \bibinfo{editor}{Fleet, D.}, \bibinfo{editor}{Pajdla, T.}, \bibinfo{editor}{Schiele, B.}, \bibinfo{editor}{Tuytelaars, T.} (Eds.), \bibinfo{booktitle}{Computer Vision -- ECCV 2014}, \bibinfo{publisher}{Springer International Publishing}, \bibinfo{address}{Cham}. pp. \bibinfo{pages}{740--755}.
\bibitem[{Luo et~al.(2021)Luo, Yang, Li and Wu}]{lue2021fpcb}
\bibinfo{author}{Luo, J.}, \bibinfo{author}{Yang, Z.}, \bibinfo{author}{Li, S.}, \bibinfo{author}{Wu, Y.}, \bibinfo{year}{2021}.
\newblock \bibinfo{title}{Fpcb surface defect detection: A decoupled two-stage object detection framework}.
\newblock \bibinfo{journal}{IEEE Transactions on Instrumentation and Measurement} \bibinfo{volume}{70}, \bibinfo{pages}{1--11}.
\newblock \DOIprefix\doi{10.1109/TIM.2021.3092510}.
\bibitem[{Luo et~al.(2023)Luo, Lin, Hao, Shen, Wu, Wang, Ruan and Zhou}]{lou2023incidence}
\bibinfo{author}{Luo, M.}, \bibinfo{author}{Lin, X.}, \bibinfo{author}{Hao, D.}, \bibinfo{author}{Shen, K.W.}, \bibinfo{author}{Wu, W.}, \bibinfo{author}{Wang, L.}, \bibinfo{author}{Ruan, S.}, \bibinfo{author}{Zhou, J.}, \bibinfo{year}{2023}.
\newblock \bibinfo{title}{Incidence and risk factors of lymph node metastasis in breast cancer patients without preoperative chemoradiotherapy and neoadjuvant therapy: analysis of seer data}.
\newblock \bibinfo{journal}{Gland Surgery} \bibinfo{volume}{12}.
\newblock \URLprefix \url{https://gs.amegroups.org/article/view/119205}.
\bibitem[{Mandelbrot(1982)}]{mandelbrot1982fractal}
\bibinfo{author}{Mandelbrot, B.B.}, \bibinfo{year}{1982}.
\newblock \bibinfo{title}{The fractal geometry of nature. 1982}.
\newblock \bibinfo{journal}{San Francisco, CA} .
\bibitem[{Martin et~al.(2001)Martin, Fowlkes, Tal and Malik}]{martin2001database}
\bibinfo{author}{Martin, D.}, \bibinfo{author}{Fowlkes, C.}, \bibinfo{author}{Tal, D.}, \bibinfo{author}{Malik, J.}, \bibinfo{year}{2001}.
\newblock \bibinfo{title}{A database of human segmented natural images and its application to evaluating segmentation algorithms and measuring ecological statistics}, in: \bibinfo{booktitle}{Proceedings Eighth IEEE International Conference on Computer Vision. ICCV 2001}, pp. \bibinfo{pages}{416--423 vol.2}.
\newblock \DOIprefix\doi{10.1109/ICCV.2001.937655}.
\bibitem[{McDonough~Dolmaya(2021)}]{dolmaya2021ethics}
\bibinfo{author}{McDonough~Dolmaya, J.}, \bibinfo{year}{2021}.
\newblock \bibinfo{title}{The ethics of crowdsourcing}.
\newblock \bibinfo{journal}{Linguistica Antverpiensia, New Series - Themes in Translation Studies} \bibinfo{volume}{10}.
\newblock \URLprefix \url{https://lans-tts.uantwerpen.be/index.php/LANS-TTS/article/view/279}, \DOIprefix\doi{10.52034/lanstts.v10i.279}.
\bibitem[{Menze et~al.(2015)Menze, Jakab, Bauer, Kalpathy-Cramer, Farahani, Kirby, Burren, Porz, Slotboom, Wiest, Lanczi, Gerstner, Weber, Arbel, Avants, Ayache, Buendia, Collins, Cordier, Corso, Criminisi, Das, Delingette, Demiralp, Durst, Dojat, Doyle, Festa, Forbes, Geremia, Glocker, Golland, Guo, Hamamci, Iftekharuddin, Jena, John, Konukoglu, Lashkari, Mariz, Meier, Pereira, Precup, Price, Raviv, Reza, Ryan, Sarikaya, Schwartz, Shin, Shotton, Silva, Sousa, Subbanna, Szekely, Taylor, Thomas, Tustison, Unal, Vasseur, Wintermark, Ye, Zhao, Zhao, Zikic, Prastawa, Reyes and Van~Leemput}]{menze2015multimodal}
\bibinfo{author}{Menze, B.H.}, \bibinfo{author}{Jakab, A.}, \bibinfo{author}{Bauer, S.}, \bibinfo{author}{Kalpathy-Cramer, J.}, \bibinfo{author}{Farahani, K.}, \bibinfo{author}{Kirby, J.}, \bibinfo{author}{Burren, Y.}, \bibinfo{author}{Porz, N.}, \bibinfo{author}{Slotboom, J.}, \bibinfo{author}{Wiest, R.}, \bibinfo{author}{Lanczi, L.}, \bibinfo{author}{Gerstner, E.}, \bibinfo{author}{Weber, M.A.}, \bibinfo{author}{Arbel, T.}, \bibinfo{author}{Avants, B.B.}, \bibinfo{author}{Ayache, N.}, \bibinfo{author}{Buendia, P.}, \bibinfo{author}{Collins, D.L.}, \bibinfo{author}{Cordier, N.}, \bibinfo{author}{Corso, J.J.}, \bibinfo{author}{Criminisi, A.}, \bibinfo{author}{Das, T.}, \bibinfo{author}{Delingette, H.}, \bibinfo{author}{Demiralp, C.}, \bibinfo{author}{Durst, C.R.}, \bibinfo{author}{Dojat, M.}, \bibinfo{author}{Doyle, S.}, \bibinfo{author}{Festa, J.}, \bibinfo{author}{Forbes, F.}, \bibinfo{author}{Geremia, E.}, \bibinfo{author}{Glocker, B.}, \bibinfo{author}{Golland, P.}, \bibinfo{author}{Guo, X.},
  \bibinfo{author}{Hamamci, A.}, \bibinfo{author}{Iftekharuddin, K.M.}, \bibinfo{author}{Jena, R.}, \bibinfo{author}{John, N.M.}, \bibinfo{author}{Konukoglu, E.}, \bibinfo{author}{Lashkari, D.}, \bibinfo{author}{Mariz, J.A.}, \bibinfo{author}{Meier, R.}, \bibinfo{author}{Pereira, S.}, \bibinfo{author}{Precup, D.}, \bibinfo{author}{Price, S.J.}, \bibinfo{author}{Raviv, T.R.}, \bibinfo{author}{Reza, S.M.S.}, \bibinfo{author}{Ryan, M.}, \bibinfo{author}{Sarikaya, D.}, \bibinfo{author}{Schwartz, L.}, \bibinfo{author}{Shin, H.C.}, \bibinfo{author}{Shotton, J.}, \bibinfo{author}{Silva, C.A.}, \bibinfo{author}{Sousa, N.}, \bibinfo{author}{Subbanna, N.K.}, \bibinfo{author}{Szekely, G.}, \bibinfo{author}{Taylor, T.J.}, \bibinfo{author}{Thomas, O.M.}, \bibinfo{author}{Tustison, N.J.}, \bibinfo{author}{Unal, G.}, \bibinfo{author}{Vasseur, F.}, \bibinfo{author}{Wintermark, M.}, \bibinfo{author}{Ye, D.H.}, \bibinfo{author}{Zhao, L.}, \bibinfo{author}{Zhao, B.}, \bibinfo{author}{Zikic, D.}, \bibinfo{author}{Prastawa, M.},
  \bibinfo{author}{Reyes, M.}, \bibinfo{author}{Van~Leemput, K.}, \bibinfo{year}{2015}.
\newblock \bibinfo{title}{The multimodal brain tumor image segmentation benchmark (brats)}.
\newblock \bibinfo{journal}{IEEE Transactions on Medical Imaging} \bibinfo{volume}{34}, \bibinfo{pages}{1993--2024}.
\newblock \DOIprefix\doi{10.1109/TMI.2014.2377694}.
\bibitem[{{Microsoft}(2024)}]{vott}
\bibinfo{author}{{Microsoft}}, \bibinfo{year}{2024}.
\newblock \bibinfo{title}{Vott}.
\newblock \bibinfo{howpublished}{\url{https://github.com/microsoft/VoTT}}.
\newblock \bibinfo{note}{[Online; accessed 01-June-2024]}.
\bibitem[{Moigne et~al.(2011)Moigne, Netanyahu and Eastman}]{moigne2011image}
\bibinfo{author}{Moigne, J.}, \bibinfo{author}{Netanyahu, N.}, \bibinfo{author}{Eastman, R.}, \bibinfo{year}{2011}.
\newblock \bibinfo{title}{Image Registration for Remote Sensing}.
\newblock \bibinfo{publisher}{Cambridge University Press}.
\newblock \URLprefix \url{https://books.google.com/books?id=v66SjxzcwIgC}.
\bibitem[{Moss et~al.(2023)Moss, Rosenzweig, Robinson, Jaffe and Litman}]{moss2023ethical}
\bibinfo{author}{Moss, A.J.}, \bibinfo{author}{Rosenzweig, C.}, \bibinfo{author}{Robinson, J.}, \bibinfo{author}{Jaffe, S.N.}, \bibinfo{author}{Litman, L.}, \bibinfo{year}{2023}.
\newblock \bibinfo{title}{Is it ethical to use mechanical turk for behavioral research? relevant data from a representative survey of mturk participants and wages}.
\newblock \bibinfo{journal}{Behavior Research Methods} \bibinfo{volume}{55}, \bibinfo{pages}{4048--4067}.
\newblock \URLprefix \url{https://doi.org/10.3758/s13428-022-02005-0}, \DOIprefix\doi{10.3758/s13428-022-02005-0}.
\bibitem[{Nickerson(1998)}]{nickerson1998confirmation}
\bibinfo{author}{Nickerson, R.S.}, \bibinfo{year}{1998}.
\newblock \bibinfo{title}{Confirmation bias: A ubiquitous phenomenon in many guises}.
\newblock \bibinfo{journal}{Review of General Psychology} \bibinfo{volume}{2}, \bibinfo{pages}{175--220}.
\newblock \URLprefix \url{https://doi.org/10.1037/1089-2680.2.2.175}, \DOIprefix\doi{10.1037/1089-2680.2.2.175}, \href{http://arxiv.org/abs/https://doi.org/10.1037/1089-2680.2.2.175}{{\tt arXiv:https://doi.org/10.1037/1089-2680.2.2.175}}.
\bibitem[{Papadopoulos et~al.(2017)Papadopoulos, Uijlings, Keller and Ferrari}]{papadopoulos2017extreme}
\bibinfo{author}{Papadopoulos, D.P.}, \bibinfo{author}{Uijlings, J.R.R.}, \bibinfo{author}{Keller, F.}, \bibinfo{author}{Ferrari, V.}, \bibinfo{year}{2017}.
\newblock \bibinfo{title}{Extreme clicking for efficient object annotation}, in: \bibinfo{booktitle}{2017 IEEE International Conference on Computer Vision (ICCV)}, pp. \bibinfo{pages}{4940--4949}.
\newblock \DOIprefix\doi{10.1109/ICCV.2017.528}.
\bibitem[{Rosenthal and Jacobson(1968)}]{rosenthal1968pygmalion}
\bibinfo{author}{Rosenthal, R.}, \bibinfo{author}{Jacobson, L.}, \bibinfo{year}{1968}.
\newblock \bibinfo{title}{Pygmalion in the classroom}.
\newblock \bibinfo{journal}{The urban review} \bibinfo{volume}{3}, \bibinfo{pages}{16--20}.
\newblock \URLprefix \url{https://doi.org/10.1007/BF02322211}, \DOIprefix\doi{10.1007/BF02322211}.
\bibitem[{Rother et~al.(2004)Rother, Kolmogorov and Blake}]{rother2004grabcut}
\bibinfo{author}{Rother, C.}, \bibinfo{author}{Kolmogorov, V.}, \bibinfo{author}{Blake, A.}, \bibinfo{year}{2004}.
\newblock \bibinfo{title}{"grabcut": Interactive foreground extraction using iterated graph cuts}.
\newblock \bibinfo{journal}{ACM Trans. Graph.} \bibinfo{volume}{23}, \bibinfo{pages}{309–314}.
\newblock \URLprefix \url{https://doi.org/10.1145/1015706.1015720}, \DOIprefix\doi{10.1145/1015706.1015720}.
\bibitem[{Roy and Haigh(2010)}]{roy2010solar}
\bibinfo{author}{Roy, I.}, \bibinfo{author}{Haigh, J.D.}, \bibinfo{year}{2010}.
\newblock \bibinfo{title}{Solar cycle signals in sea level pressure and sea surface temperature}.
\newblock \bibinfo{journal}{Atmospheric Chemistry and Physics} \bibinfo{volume}{10}, \bibinfo{pages}{3147--3153}.
\newblock \URLprefix \url{https://acp.copernicus.org/articles/10/3147/2010/}, \DOIprefix\doi{10.5194/acp-10-3147-2010}.
\bibitem[{Russakovsky et~al.(2015)Russakovsky, Deng, Su, Krause, Satheesh, Ma, Huang, Karpathy, Khosla, Bernstein et~al.}]{russakovsky2015imagenet}
\bibinfo{author}{Russakovsky, O.}, \bibinfo{author}{Deng, J.}, \bibinfo{author}{Su, H.}, \bibinfo{author}{Krause, J.}, \bibinfo{author}{Satheesh, S.}, \bibinfo{author}{Ma, S.}, \bibinfo{author}{Huang, Z.}, \bibinfo{author}{Karpathy, A.}, \bibinfo{author}{Khosla, A.}, \bibinfo{author}{Bernstein, M.}, et~al., \bibinfo{year}{2015}.
\newblock \bibinfo{title}{Imagenet large scale visual recognition challenge}.
\newblock \bibinfo{journal}{International journal of computer vision} \bibinfo{volume}{115}, \bibinfo{pages}{211--252}.
\newblock \DOIprefix\doi{https://doi.org/10.1007/s11263-015-0816-y}.
\bibitem[{Russell et~al.(2008)Russell, Torralba, Murphy and Freeman}]{russell2008labelme}
\bibinfo{author}{Russell, B.C.}, \bibinfo{author}{Torralba, A.}, \bibinfo{author}{Murphy, K.P.}, \bibinfo{author}{Freeman, W.T.}, \bibinfo{year}{2008}.
\newblock \bibinfo{title}{Labelme: a database and web-based tool for image annotation}.
\newblock \bibinfo{journal}{International journal of computer vision} \bibinfo{volume}{77}, \bibinfo{pages}{157--173}.
\newblock \DOIprefix\doi{https://doi.org/10.1007/s11263-007-0090-8}.
\bibitem[{Sishodia et~al.(2020)Sishodia, Ray and Singh}]{sishodia2020applications}
\bibinfo{author}{Sishodia, R.P.}, \bibinfo{author}{Ray, R.L.}, \bibinfo{author}{Singh, S.K.}, \bibinfo{year}{2020}.
\newblock \bibinfo{title}{Applications of remote sensing in precision agriculture: A review}.
\newblock \bibinfo{journal}{Remote Sensing} \bibinfo{volume}{12}.
\newblock \URLprefix \url{https://www.mdpi.com/2072-4292/12/19/3136}, \DOIprefix\doi{10.3390/rs12193136}.
\bibitem[{Standing and Standing(2018)}]{standing2018ethical}
\bibinfo{author}{Standing, S.}, \bibinfo{author}{Standing, C.}, \bibinfo{year}{2018}.
\newblock \bibinfo{title}{The ethical use of crowdsourcing}.
\newblock \bibinfo{journal}{Business Ethics: A European Review} \bibinfo{volume}{27}, \bibinfo{pages}{72--80}.
\newblock \URLprefix \url{https://onlinelibrary.wiley.com/doi/abs/10.1111/beer.12173}, \DOIprefix\doi{https://doi.org/10.1111/beer.12173}, \href{http://arxiv.org/abs/https://onlinelibrary.wiley.com/doi/pdf/10.1111/beer.12173}{{\tt arXiv:https://onlinelibrary.wiley.com/doi/pdf/10.1111/beer.12173}}.
\bibitem[{Su et~al.(2012)Su, Deng and Fei-Fei}]{su2012crowdsourcing}
\bibinfo{author}{Su, H.}, \bibinfo{author}{Deng, J.}, \bibinfo{author}{Fei-Fei, L.}, \bibinfo{year}{2012}.
\newblock \bibinfo{title}{Crowdsourcing annotations for visual object detection}, in: \bibinfo{booktitle}{Human Computation - Papers from the 2012 AAAI Workshop, Technical Report}, pp. \bibinfo{pages}{40--46}.
\newblock \bibinfo{note}{2012 AAAI Workshop ; Conference date: 23-07-2012 Through 23-07-2012}.
\bibitem[{{Team}(2022a)}]{citizenscience2022JetHunter}
\bibinfo{author}{{Team}}, \bibinfo{year}{2022}a.
\newblock \bibinfo{title}{Citizen science - solar jet hunter}.
\newblock \bibinfo{howpublished}{\url{https://www.zooniverse.org/projects/sophiemu/solar-jet-hunter}}.
\newblock \bibinfo{note}{[Online; accessed 13-June-2023]}.
\bibitem[{{Team}(2022b)}]{openimagedataset2022google}
\bibinfo{author}{{Team}}, \bibinfo{year}{2022}b.
\newblock \bibinfo{title}{Open images dataset website}.
\newblock \bibinfo{howpublished}{\url{https://storage.googleapis.com/openimages/web/index.html}}.
\newblock \bibinfo{note}{[Online; accessed 13-June-2023]}.
\bibitem[{Team(2024a)}]{cvat}
\bibinfo{author}{Team}, \bibinfo{year}{2024}a.
\newblock \bibinfo{title}{Cvat}.
\newblock \bibinfo{howpublished}{\url{https://www.cvat.ai/}}.
\newblock \bibinfo{note}{Online; accessed 01-June-2024}.
\bibitem[{Team(2024b)}]{dataloop}
\bibinfo{author}{Team}, \bibinfo{year}{2024}b.
\newblock \bibinfo{title}{dataloop}.
\newblock \bibinfo{howpublished}{\url{https://dataloop.ai/}}.
\newblock \bibinfo{note}{Online; accessed 01-June-2024}.
\bibitem[{{Team}(2024)}]{labelstudio}
\bibinfo{author}{{Team}}, \bibinfo{year}{2024}.
\newblock \bibinfo{title}{Label studio}.
\newblock \bibinfo{howpublished}{\url{https://labelstud.io/}}.
\newblock \bibinfo{note}{[Online; accessed 01-June-2024]}.
\bibitem[{Team(2024)}]{labelbox}
\bibinfo{author}{Team}, \bibinfo{year}{2024}.
\newblock \bibinfo{title}{Labelbox}.
\newblock \bibinfo{howpublished}{\url{https://labelbox.com/}}.
\newblock \bibinfo{note}{Online; accessed 01-June-2024}.
\bibitem[{{Team}(2024)}]{labelimg}
\bibinfo{author}{{Team}}, \bibinfo{year}{2024}.
\newblock \bibinfo{title}{Labelimg}.
\newblock \bibinfo{howpublished}{\url{https://github.com/HumanSignal/labelImg}}.
\newblock \bibinfo{note}{[Online; accessed 01-June-2024]}.
\bibitem[{Team(2024a)}]{scaleAI}
\bibinfo{author}{Team}, \bibinfo{year}{2024}a.
\newblock \bibinfo{title}{Scale ai}.
\newblock \bibinfo{howpublished}{\url{https://scale.com/data-engine}}.
\newblock \bibinfo{note}{Online; accessed 01-June-2024}.
\bibitem[{Team(2024b)}]{superAnnotate}
\bibinfo{author}{Team}, \bibinfo{year}{2024}b.
\newblock \bibinfo{title}{Superannotate}.
\newblock \bibinfo{howpublished}{\url{https://www.superannotate.com/}}.
\newblock \bibinfo{note}{Online; accessed 01-June-2024}.
\bibitem[{Team(2024c)}]{V7}
\bibinfo{author}{Team}, \bibinfo{year}{2024}c.
\newblock \bibinfo{title}{V7}.
\newblock \bibinfo{howpublished}{\url{https://www.v7labs.com/}}.
\newblock \bibinfo{note}{Online; accessed 01-June-2024}.
\bibitem[{Tejani et~al.(2024)Tejani, Ng, Xi and Rayan}]{tejani2024understanding}
\bibinfo{author}{Tejani, A.S.}, \bibinfo{author}{Ng, Y.S.}, \bibinfo{author}{Xi, Y.}, \bibinfo{author}{Rayan, J.C.}, \bibinfo{year}{2024}.
\newblock \bibinfo{title}{Understanding and mitigating bias in imaging artificial intelligence}.
\newblock \bibinfo{journal}{RadioGraphics} \bibinfo{volume}{44}, \bibinfo{pages}{e230067}.
\newblock \URLprefix \url{https://doi.org/10.1148/rg.230067}, \DOIprefix\doi{10.1148/rg.230067}, \href{http://arxiv.org/abs/https://doi.org/10.1148/rg.230067}{{\tt arXiv:https://doi.org/10.1148/rg.230067}}. \bibinfo{note}{pMID: 38635456}.
\bibitem[{{Theiler}(1990)}]{theiler1990estimating}
\bibinfo{author}{{Theiler}, J.}, \bibinfo{year}{1990}.
\newblock \bibinfo{title}{Estimating fractal dimension}.
\newblock \bibinfo{journal}{Journal of The Optical Society of America A-optics Image Science and Vision} \bibinfo{volume}{7}, \bibinfo{pages}{1055--1073}.
\newblock \URLprefix \url{https://doi.org/10.1364/JOSAA.7.001055}, \DOIprefix\doi{10.1364/JOSAA.7.001055}.
\bibitem[{Tsukiyama et~al.(1986)Tsukiyama, Kondo, Kakuse, Saba, Ozaki and Itoh}]{tsukiyama1986method}
\bibinfo{author}{Tsukiyama, T.}, \bibinfo{author}{Kondo, Y.}, \bibinfo{author}{Kakuse, K.}, \bibinfo{author}{Saba, S.}, \bibinfo{author}{Ozaki, S.}, \bibinfo{author}{Itoh, K.}, \bibinfo{year}{1986}.
\newblock \bibinfo{title}{Method and system for data compression and restoration}.
\newblock \bibinfo{note}{US Patent 4,586,027}.
\bibitem[{Velasco(2024)}]{Velasco_2024}
\bibinfo{author}{Velasco, L.}, \bibinfo{year}{2024}.
\newblock \bibinfo{title}{The hidden labor force behind chatgpt: The drama of the “ghost workers”}.
\newblock \URLprefix \url{https://english.elpais.com/economy-and-business/2024-01-01/the-hidden-labor-force-behind-chatgpt-the-drama-of-the-ghost-workers.html}.
\bibitem[{Vondrick et~al.(2013)Vondrick, Patterson and Ramanan}]{vondrick2013efficiently}
\bibinfo{author}{Vondrick, C.}, \bibinfo{author}{Patterson, D.}, \bibinfo{author}{Ramanan, D.}, \bibinfo{year}{2013}.
\newblock \bibinfo{title}{Efficiently scaling up crowdsourced video annotation: A set of best practices for high quality, economical video labeling}.
\newblock \bibinfo{journal}{International journal of computer vision} \bibinfo{volume}{101}, \bibinfo{pages}{184--204}.
\newblock \URLprefix \url{https://doi.org/10.1007/s11263-012-0564-1}, \DOIprefix\doi{10.1007/s11263-012-0564-1}.
\bibitem[{Wang(2017)}]{wang2017heterogenous}
\bibinfo{author}{Wang, L.}, \bibinfo{year}{2017}.
\newblock \bibinfo{title}{Heterogeneous data and big data analytics}.
\newblock \bibinfo{journal}{Automatic Control and Information Sciences} \bibinfo{volume}{3}, \bibinfo{pages}{8--15}.
\newblock \URLprefix \url{http://pubs.sciepub.com/acis/3/1/3}, \DOIprefix\doi{10.12691/acis-3-1-3}.
\bibitem[{Yu et~al.(2015)Yu, Seff, Zhang, Song, Funkhouser and Xiao}]{yu2015lsun}
\bibinfo{author}{Yu, F.}, \bibinfo{author}{Seff, A.}, \bibinfo{author}{Zhang, Y.}, \bibinfo{author}{Song, S.}, \bibinfo{author}{Funkhouser, T.}, \bibinfo{author}{Xiao, J.}, \bibinfo{year}{2015}.
\newblock \bibinfo{title}{Lsun: Construction of a large-scale image dataset using deep learning with humans in the loop}.
\newblock \bibinfo{journal}{arXiv preprint arXiv:1506.03365} .
\bibitem[{Zhou et~al.(2014)Zhou, Lapedriza, Xiao, Torralba and Oliva}]{zhou2014learning}
\bibinfo{author}{Zhou, B.}, \bibinfo{author}{Lapedriza, A.}, \bibinfo{author}{Xiao, J.}, \bibinfo{author}{Torralba, A.}, \bibinfo{author}{Oliva, A.}, \bibinfo{year}{2014}.
\newblock \bibinfo{title}{Learning deep features for scene recognition using places database}, in: \bibinfo{editor}{Ghahramani, Z.}, \bibinfo{editor}{Welling, M.}, \bibinfo{editor}{Cortes, C.}, \bibinfo{editor}{Lawrence, N.}, \bibinfo{editor}{Weinberger, K.} (Eds.), \bibinfo{booktitle}{Advances in Neural Information Processing Systems}, \bibinfo{publisher}{Curran Associates, Inc.}. p. \bibinfo{pages}{TBD}.
\newblock \URLprefix \url{https://proceedings.neurips.cc/paper_files/paper/2014/file/3fe94a002317b5f9259f82690aeea4cd-Paper.pdf}.

\end{thebibliography}

\end{document}